\pdfoutput=1
\PassOptionsToPackage{prologue,dvipsnames}{xcolor}

\documentclass[11pt]{article}

\usepackage[]{acl}

\usepackage{times}
\usepackage{latexsym}
\usepackage{amssymb}
\usepackage{amsmath}
\usepackage{amsthm}
\usepackage{booktabs}
\usepackage{enumitem}
\usepackage{graphicx}
\usepackage{color}
\usepackage{microtype}
\usepackage{mathtools}  
\usepackage{subcaption}
\usepackage{xcolor, colortbl}
\usepackage{booktabs}
\usepackage{algorithm}
\usepackage{epsfig}
\usepackage{graphicx}
\usepackage{amssymb}
\usepackage{epsfig}
\usepackage{amsmath, mdframed}
\usepackage{multirow}
\usepackage{makecell}
\usepackage{setspace}
\DeclareMathOperator*{\argmax}{arg\,max}
\newcolumntype{P}[1]{>{\centering\arraybackslash}p{#1}}
\usepackage{color,soul}
\usepackage[export]{adjustbox}
\usepackage{url}
\usepackage{pifont}
\usepackage{fancyvrb}
\usepackage[dvipsnames]{xcolor}
\usepackage{listings}
\usepackage{ulem}

\lstdefinestyle{mycustomstyle}{
    basicstyle=\ttfamily,        
    breaklines=true,             
    keepspaces=true,             
    xleftmargin=0pt,            
    xrightmargin=0pt,           
    frame=none,                  
    escapeinside={(*@}{@*)},     
}
\usepackage[T1]{fontenc}

\usepackage[utf8]{inputenc}

\usepackage{microtype}

\usepackage{inconsolata}

\usepackage{algpseudocode}
\usepackage{spverbatim}
\usepackage{hyperref}

\newcommand{\SCULPT}{SCULPT }
\newcommand{\protegi}{ProTeGi}

\newcommand{\changeprompt}[1]{\textcolor{ForestGreen}{#1}}

\newcommand{\rephrasedPrompt}[1]{\textcolor{BlueGreen}{#1}}
\newcommand{\examplePrompt}[1]{\textcolor{Mahogany}{#1}}

\title{SCULPT: Systematic Tuning of Long Prompts}

\author{Shanu Kumar \quad Akhila Yesantarao Venkata \quad Shubhanshu Khandelwal\\
\textbf{Bishal Santra \quad Parag Agrawal\quad Manish Gupta}\\
Microsoft Corporation, India \\
{\tt \small \{shankum,akyesant,shukhand,bishalsantra,paragag,gmanish\}@microsoft.com} }

\usepackage[toc,page,header]{appendix}
\usepackage{minitoc}

\begin{document}
\doparttoc 
\faketableofcontents 
\maketitle
\begin{abstract}  
Prompt optimization is essential for effective utilization of large language models (LLMs) across diverse tasks. While existing optimization methods are effective in optimizing short prompts, they struggle with longer, more complex ones, often risking information loss and being sensitive to small perturbations. To address these challenges, we propose \textbf{SCULPT} (\textit{\textbf{S}ystemati\textbf{c} T\textbf{u}ning of \textbf{L}ong \textbf{P}romp\textbf{t}s}), a framework that treats prompt optimization as a hierarchical tree refinement problem. SCULPT represents prompts as tree structures, enabling targeted modifications while preserving contextual integrity. It employs a \textit{Critic-Actor} framework that generates reflections and applies actions to refine the prompt. Evaluations demonstrate SCULPT’s effectiveness on long prompts, its robustness to adversarial perturbations, and its ability to generate high-performing prompts even without any initial human-written prompt. Compared to existing state of the art methods, SCULPT consistently improves LLM performance by preserving essential task information while applying structured refinements. Both qualitative and quantitative analyses show that SCULPT produces more stable and interpretable prompt modifications, ensuring better generalization across tasks.   
\end{abstract}

\section{Introduction}  
Large language models (LLMs) have revolutionized natural language processing, achieving state-of-the-art performance in text generation, summarization, and reasoning \cite{achiam2023gpt, bubeck2023sparks, abdin2024phi, dubey2024llama}. A key factor in their success is the use of natural language prompts, which condition the model on specific tasks. As applications grow in complexity, prompts have become not only longer but also structurally intricate, often spanning hundreds or even thousands of tokens and integrating multiple instructions, examples, and contextual cues \cite{schnabel2024prompts}. Optimizing such prompts manually is time-consuming, requiring expert intervention and extensive trial-and-error iterations \cite{jiang2022promptmaker, zamfirescu2023johnny}.  

\begin{figure*}[!ht]
    \centering
    \includegraphics[scale=0.095]{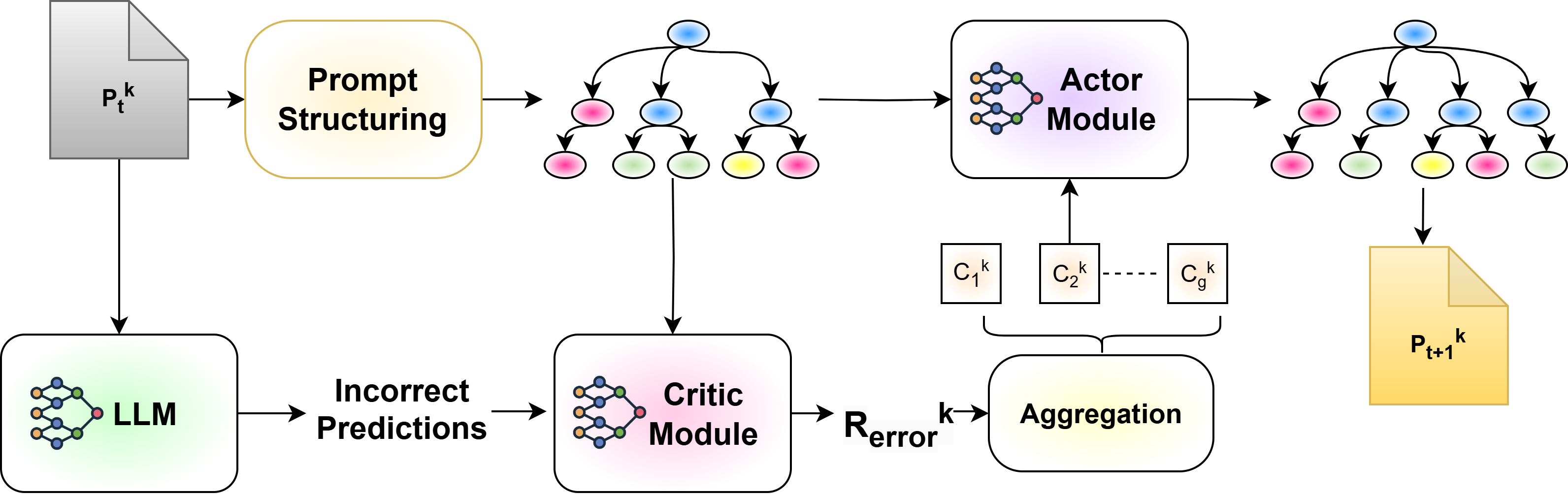}
    \caption{Overview of the SCULPT framework, highlighting its four core components: Prompt Structuring, the Critic Module, Aggregation, and the Actor Module for optimizing the $k$-th candidate prompt $\mathcal{P}_{t}^k$ at  iteration $t$. We have omitted UCB-based prompt selection and Structural Reflection in the figure for clarity.}
    \label{fig:sculpt_diagram}
    \vspace{-1em}
\end{figure*}

To reduce manual effort, automatic prompt optimization methods such as APE~\citep{zhou2022large}, \protegi~\citep{pryzant2023automatic}, OPRO~\citep{yang2023large}, and APEX~\citep{hsieh2023automatic} have been proposed. These methods have been evaluated on tasks where prompts consist of minimal instructions, demonstrating their effectiveness in optimizing short prompts. However, they face two major limitations when applied to longer prompts. First, they generate each token of new prompt candidates from scratch, risking the loss of information from the initial prompt. Second, due to the non-convex and non-monotonic behavior of LLMs with respect to small perturbations in prompt structure \cite{jiang-etal-2020-know, zhao2021calibrate, reynolds2021prompt, lu-etal-2022-fantastically}, these optimization techniques become ineffective for long prompts. Addressing these limitations requires a structured and context-aware approach that preserves the initial information while applying targeted refinements.

We address these challenges with \textbf{SCULPT} (\textit{\textbf{S}ystemati\textbf{c} T\textbf{u}ning of \textbf{L}ong \textbf{P}romp\textbf{t}s})\footnote{Our code is available at \url{https://github.com/Sshanu/SCULPT}}, a framework that redefines prompt optimization as a hierarchical tree refinement problem. Rather than treating prompts as flat sequences, SCULPT represents a prompt as a tree-structured form. This representation retains the intrinsic structure of a long prompt while enabling targeted and effective modifications. SCULPT employs an iterative \textit{Critic-Actor} framework: the \textit{Critic Module} generates reflections based on the prompt tree and incorrect predictions, while the \textit{Actor Module} processes these reflections and generates a list of actions inspired by expert-driven prompt optimization. These actions are then applied systematically to refine the prompt tree. Fig.~\ref{fig:sculpt_diagram} provides an overview of our proposed SCULPT framework.

Our contributions are as follows:  
(1) We introduce \textit{SCULPT}, a novel framework for optimizing long prompts using a hierarchical tree structure and an actor-critic mechanism, enabling systematic and targeted refinements.  
(2) We demonstrate \textit{SCULPT}’s effectiveness in refining unstructured prompts, achieving significant gains in LLM performance across four BBH (Big Bench Hard) tasks, four RAI (Responsible AI) tasks, and two multi-label tasks, with initial prompts averaging 1000 words and a maximum length of 2,644 words.  
(3) We evaluate \textit{SCULPT} in adversarial and autogenerated prompt settings, showing its ability to refine perturbed prompts and generate effective prompts without human-curated initial prompts.  
(4) We analyze structural and semantic differences using three metrics, demonstrating \textit{SCULPT}’s ability to refine prompts while preserving key information.
(5) We assess \textit{SCULPT}’s action distribution, demonstrating its controlled, systematic, and balanced refinements, leading to stable and generalizable prompt optimizations.  

\section{Related Work}
\label{sec:related_work}

Optimizing prompts is essential for maximizing LLM performance across various tasks \cite{brown2020language, reynolds2021prompt, wang2022self, chang2024efficient, sahoo2024systematic}. While manual prompt engineering has been effective, it is labor-intensive and requires expertise. To automate this process, \textit{soft prompting} methods \cite{lester2021power, li2021prefix, liu2021gpt, qin2021learning} optimize prompts as continuous vectors in the model's embedding space, but they require access to model weights, making them unsuitable for black-box LLMs. In contrast, \textit{black-box prompt optimization} techniques refine prompts without modifying the internal model, relying on explicit or implicit reflection mechanisms.  

Explicit reflection-based approaches \cite{cheng2023black, pryzant2023automatic, ye2023prompt, sun2023autohint, dong-etal-2024-pace} generate feedback based on model errors and iteratively refine prompts by incorporating this feedback. We adopt this approach by structuring reflections to optimize long prompts effectively. In contrast, implicit reflection-based methods, such as OPRO \cite{yang2023large} and evolutionary algorithms \cite{xu2022gps, guo2024connecting, pmlr-v235-fernando24a, liu2024large}, improve prompts using historical performance rather than explicit feedback. Some methods further incorporate human preferences to enhance the optimization efficiency \cite{chen2024prompt}. These techniques have also been integrated into multi-step AI pipelines to improve their prompt quality \cite{khattab2023dspy, yuksekgonul2024textgrad, schnabel2024prompts}. Additionally, research on automatic prompt generation explores approaches that construct prompts from input-output pairs \cite{honovich-etal-2023-instruction, zhou2022large, pmlr-v235-chen24e}.

Recent studies have explored prompt optimization for longer prompts by applying segmentation and predefined modifications \cite{prasad2022grips}. However, these methods remain limited in scope. APEX \cite{hsieh2023automatic}, for instance, optimizes few-shot chain-of-thought prompts but struggles with complex, instruction-heavy prompts. Additionally, many existing optimization techniques exhibit unpredictable behavior, leading to suboptimal results \cite{ma2024large}. To address these challenges, we introduce targeted updates that ensure stable and controlled refinements of long prompts.

\begin{figure*}[!ht]
    \centering
    \includegraphics[scale=0.075]{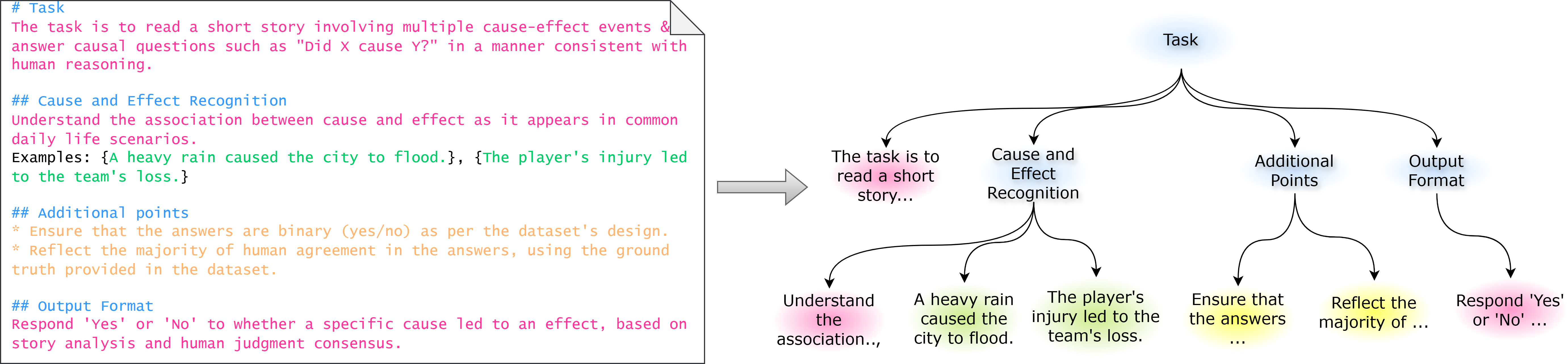}
    \caption{Illustration of SCULPT's Prompt Structuring Process. Unstructured prompt is transformed into a hierarchical tree structure, with different colors represent various node types (e.g., heading, instructions, examples).}
    \label{fig:prompt_structure}
    \vspace{-1em}
\end{figure*}

\section{The SCULPT Methodology}
\label{sec:methodology}

In this section, we present SCULPT, a framework designed to optimize complex, long prompts for LLMs. While existing methods primarily focus on short prompts or few-shot examples, \SCULPT specifically addresses the challenges of optimizing longer prompts containing multiple instructions, examples, and layered structures. Our goal is to refine prompts iteratively in a controlled manner, ensuring robust model performance while maintaining clarity and task relevance. Let \(\mathcal{D}_{\text{train}}, \mathcal{D}_{\text{val}}, \mathcal{D}_{\text{test}}\) represent the training, validation, and test datasets, each consisting of input-output pairs \((x, y)\). The LLM \(\mathcal{M}\) generates predictions \(\hat{y} = \mathcal{M}(\mathcal{P}, x)\) based on the given prompt \(\mathcal{P}\), which can contain complex instructions and examples (Appendix \ref{sec:initial_prompts}). The objective of \SCULPT is to find an optimized prompt \(\mathcal{P}^*\) that maximizes a performance metric \(\mathcal{Q}\) (e.g., accuracy) on \(\mathcal{D}_{\text{val}}\):
\vspace{-0.5em}
\[
\mathcal{P}^* = \arg\max_{\mathcal{P}} \mathbb{E}_{(x, y) \sim \mathcal{D}_{\text{val}}} \left[ \mathcal{Q}(y, \mathcal{M}(\mathcal{P}, x)) \right]
\]
Starting with an initial prompt $P_{t=0}$, the optimization process iteratively produces $K$ candidate prompts $\{P^k_t\}_{k=1}^K$ at every iteration $t$.
\SCULPT consists of four core components: \textit{Prompt Structuring}, \textit{Critic Module}, \textit{Aggregation of Reflections}, and \textit{Actor Module}, working in conjunction with a beam search strategy to explore and optimize multiple candidate prompts simultaneously. As illustrated in Fig.~\ref{fig:sculpt_diagram}, these components systematically refine prompts by structuring, analyzing, aggregating, and applying controlled modifications. Henceforth, for sake of clarity, we will drop subscript $t$.

\subsection{Prompt Structuring}  
Short prompt optimization methods struggle with longer, more complex instructions, making it difficult to attribute error feedback to specific sections. Treating a long prompt as a single unit often leads to fragmented and ineffective refinements. To address this, we represent prompts as a hierarchical tree \(\mathcal{T} = (N, E)\), where \(N\) is the set of nodes representing components such as headings, instructions, and examples, while \(E\) defines containment relationships between nodes. This structure enables targeted modifications while preserving the integrity of unrelated sections.

Given a prompt \(\mathcal{P}^k\), it is transformed into its hierarchical representation \(\mathcal{T}^k\). If the prompt has an explicit structure, such as markdown formatting, it is directly parsed into \(\mathcal{T}^k\); otherwise, an LLM infers the hierarchy by segmenting the prompt into distinct components while preserving logical relationships (Appendix~\ref{app:structuringPrompt}). This enables \SCULPT to effectively process prompts with any type of formatting. Fig.~\ref{fig:prompt_structure} illustrates this transformation, with different node types color-coded to represent the hierarchical structure.

\subsection{Critic Module}
The \textit{Critic Module} \(\mathbb{C}\) evaluates the prompt and generates two types of reflections: \textit{Structural Reflection} and \textit{Error Reflection}. Each reflection includes feedback and a list of paths to the nodes in \(\mathcal{T}^{k}\) where modifications should be applied. \textit{Structural Reflection} (\(\mathcal{R}_{\text{struc}}\)) assesses the overall structure, clarity, completeness, and redundancy of the prompt. It is generated independently of task-specific errors and ensures logical organization, given by \(\mathcal{R}^k_{\text{struc}} = \mathbb{C}(\mathcal{T}^{k})\).  

\textit{Error Reflection} (\(\mathcal{R}_{\text{error}}\)) is generated when \(\hat{y} \neq y\) for an input-output pair in the training batch \(B \subset \mathcal{D}_{\text{train}}\), identifying problematic nodes in \(\mathcal{T}^{k}\) that contribute to incorrect predictions, formulated as \(\mathcal{R}^k_{\text{error}} = \bigl\{ \mathbb{C}(\mathcal{T}^{k}, x_i, y_i, \hat{y}_i) : i \in B \bigr\}\). Since error reflections are highly specific to individual examples, using them directly may lead to overfitting. To enhance generalization, \SCULPT aggregates these reflections before applying modifications.

\subsection{Aggregation of Reflections}
To mitigate overfitting, \SCULPT consolidates error reflections \(\mathcal{R}^k_{\text{error}}\) into a structured set \(\mathbf{C}^k_{\text{error}} = \{C^k_1, C^k_2, \dots, C^k_g\}\), where \(g\) is determined by the aggregation mechanism. We employ two complementary strategies: \textit{Pattern-based Aggregation}, which clusters reflections based on shared error types and structural similarities, and \textit{Node-based Aggregation}, which groups reflections corresponding to the same node. If a node \( N_j \) appears in at least one reflection in \( \mathcal{R}^k_{\text{error}} \), its aggregated reflection \( C^k_j \) is defined as:
\vspace{-0.7em}
\[
C^k_j = \bigcup \{ \mathcal{R} \mid \mathcal{R} \in \mathcal{R}^k_{\text{error}} \text{ and } N_j \in \mathcal{R} \}
\]

This ensures that all reflections affecting the same node are merged, allowing for more structured and meaningful modifications.

\subsection{Actor Module}
The \textit{Actor Module} modifies the prompt based on reflections from the Critic Module. Instead of operating on the entire prompt tree \(\mathcal{T}^{k}\), the Actor focuses on an induced subtree \(\mathcal{T}_{\text{sub}}^{k}\), which includes nodes requiring modification along with their direct parent nodes. Given \(\mathcal{T}_{\text{sub}}^{k}\) and any reflection, the Actor generates a list of actions \( A^k = \{a_1, a_2, \dots, a_m\} \) selected from a predefined set of modifications outlined in Table \ref{tab:actions}. These actions are then applied using an update operator \(\Phi\), transforming the subtree into an updated version:
\vspace{-0.7em}
\[
\mathcal{T}^{k}_{\text{sub,} t+1} = \Phi\Big(\mathcal{T}_{\text{sub},t}^{k}, A^k\Big)
\]

The Actor first applies high-level structural modifications derived from \(\mathcal{R}_{\text{struc}}^k\) to improve clarity and logical organization. It then incorporates aggregated reflections \(C_j^k\) to refine instructions and examples based on task-specific errors. Once all modifications are completed, the updated prompt tree \(\mathcal{T}^k_{t+1}\) is converted back into its textual representation, yielding the optimized prompt \(\mathcal{P}^k_{t+1}\).

\begin{table}[h!]
\scriptsize
\centering
\begin{tabular}{ll}
\toprule
\textbf{Action}             & \textbf{Description}  \\
\midrule
\textit{Structural Reordering} & Changing the order of sibling nodes \\ 
\textit{Instruction Update}   & Simplifying or adding new instructions \\ 
\textit{Example Addition}     & Adding new examples to a node  \\ 
\textit{Example Deletion}     & Removing redundant examples from a node \\
\textit{Example Refinement}   & Improving existing examples in a node \\ 
\textit{Node Pruning}         & Removing unnecessary nodes  \\ 
\textit{Node Expansion}       & Adding new nodes to address gaps \\ 
\textit{Node Merging}         & Combining nodes that have similar content \\
\bottomrule
\end{tabular}
\vspace{-1em}
\caption{Action types in \SCULPT for prompt refinement}
\label{tab:actions}
\end{table}

\begin{algorithm}[h!]
\scriptsize
\caption{Prompt Optimization in \SCULPT$^1$}
\begin{algorithmic}
\State \textbf{Initialize} Beam $\mathcal{B}_0 = \{ \mathcal{P}_0 \}$, $t \gets 0$, max\_steps
\While{$t < $ max\_steps} 
    \State Evaluate $\mathcal{B}_t$ on a random subset of $\mathcal{D}_{\text{val}}$, obtain $\hat{\mu}_k$
    \State Compute UCB scores $\text{UCB}_k(t)$ for each $\mathcal{P}_t^k$
    \State Select top $K$ candidate prompts $\{\mathcal{P}_t^k\}_{k=1}^K$
    \For{each selected candidate prompt $\mathcal{P}_t^k$}
        \State Critic generates \textit{Structural Reflection} $\mathcal{R}_{\text{struc}}^k$
        \State Critic generates \textit{Error Reflection} $\mathcal{R}_{\text{error}}^k$
        \State Aggregate $\mathcal{R}_{\text{error}}^k$ into groups $\{ \mathcal{C}_j^k \}_{j=1}^g$
        \State Actor applies structural actions ${A}_{\text{struc}}^k$
        \For{each aggregated reflection $ \mathcal{C}_j^k$}
            \State Actor applies error-based actions ${A}_{\text{error}}^{(k, j)}$
            \State Update prompt to $\mathcal{P}_{t+1}^{(k, j)}$ and add to beam $\mathcal{B}_{t+1}$ 
        \EndFor
    \EndFor
    \State $t \gets t + 1$
\EndWhile
\State \textbf{Return} top-$K$ prompts from $\mathcal{B}_t$ sorted in descending order by UCB scores
\end{algorithmic}
\label{algo:sculpt}
\end{algorithm}

\vspace{-1em}
\subsection{Search Process}

\SCULPT incorporates a beam search strategy to explore and refine multiple candidate prompts in parallel. At each iteration \(t\), a beam \(\mathcal{B}_t\) maintains the top \(K\) candidate prompts, enabling both exploitation of high-performing prompts and exploration of new variations. Since evaluating all candidate prompts on \(\mathcal{D}_{\text{val}}\) is computationally expensive, we adopt an Upper Confidence Bound (UCB)-based selection strategy \cite{pryzant2023automatic}. The UCB score for each candidate \(\mathcal{P}_t^k\) is computed as:
\vspace{-0.5em}
\[
\text{UCB}_k(t) = \hat{\mu}_k + c \sqrt{\frac{\log t}{n_k}}
\]

\begin{table*}[h!]
\centering
\scriptsize
\setlength\tabcolsep{9pt}
\begin{tabular}{lccccccccccc}
\toprule
\textbf{Method} & \textbf{ST} & \textbf{Dis} & \textbf{CJ} & \textbf{FF} & \textbf{Inapp} & \textbf{Misinfo} & \textbf{Hate} & \textbf{Selfharm} & \textbf{BTails} & \textbf{GoE} & \textbf{Avg} \\
\midrule
Initial Prompt  & 62.3 & 74.1 & 71.1 & 80.2 & 46.6 & 51.5 & 46.8 & 66.4 & 41.8 & 7.8  & 54.9 \\
APE             & 51.0 & 74.3 & 71.8 & 75.3 & 45.3 & 31.9 & 29.3 & 38.1 & 46.4   & 0   & 46.3 \\
LAPE        & 52.7 & 78.0 & 72.2 & 81.3 & 42.4 & 43.6 & 37.6 & 44.2 & 39.1   & 19.6   & 50.3 \\
APEX            & 62.7 & 61.5 & 70.5 & 80.6 & 48.0 & 50.5 & 47.1 & 65.2 & 40.7 & 8.0  & 53.5 \\
OPRO            & 64.6 & 75.1 & 72.7 & 80.7 & 46.6 & 51.0 & 40.9 & 65.1 & 41.5 & 11.8 & 55.0 \\
ProTeGi         & 65.5 & 74.8 & 68.9 & 70.1 & 44.8 & 54.8 & 51.9 & 66.5 & 45.4 & 17.8 & 56.1 \\
\midrule
$\text{SCULPT}_{\text{NoAgg}}$       & 65.2 & 77.3 & 75.4 & 83.1 & 53.6 & 53.6 & 51.9 & 65.5 & 49.6   & 29.8   & 60.5\\
$\text{SCULPT}_{\text{PA}}$       & 66.2 & 77.6 & \textbf{76.9 }& 83.7 & \textbf{55.3} & \textbf{56.7} & \textbf{53.1} & 68.8 & 49.3   & 29.6   & 61.7 \\
SCULPT+RP       & 67.6 & 78.0 & 75.1 & 84.7 & 55.0 & 55.3 & 52.9 & \textbf{69.0} & 49.6   & 22.4   & 61.0 \\
SCULPT          & \textbf{68.8} & {80.1} & 75.9 & 83.7 & 55.0 & 54.9 & \textbf{53.1} & 68.5 & \textbf{50.5} & \textbf{30.6} & \textbf{62.1} \\
$\text{SCULPT}_{\text{LAPE}}$ & 66.8 & \textbf{81.1} & 76.1 & \textbf{86.5} & 48.2 & 48.8 & 44.5 & 61.8 & 48.4   & 28.0   & 59.0 \\
\bottomrule
\end{tabular}
\vspace{-1em}
\caption{Performance comparison using GPT-4o across various tasks}
\label{tab:gpt_result}
\end{table*}

where \(\hat{\mu}_k\) is the estimated performance (using previous evaluations) of the candidate on the validation set \(\mathcal{D}_{\text{val}}\), \(n_k\) is the number of times the candidate has been evaluated, and \(c\) is a hyperparameter controlling the trade-off between exploration and exploitation. This ensures that promising candidates with fewer evaluations are prioritized while refining high-performing candidates.
\footnotetext[1]{We have removed $t$ from notations for $\mathcal{C}$, $A$, $\mathcal{R}$ for clarity.}

The overall optimization process for SCULPT, integrating prompt structuring, reflection-based refinement, and beam search with UCB-based selection, is provided in Algorithm \ref{algo:sculpt}. A more detailed explanation of the beam search algorithm and the UCB-based selection strategy is provided in Appendix \ref{sec:ucb}. Additionally, detailed templates for the Critic and Actor modules are presented in Appendix \ref{sec:sculpt_prompts}, while step-by-step interactions and refinements are detailed in Appendix \ref{sec:critic_actor_response}. Fig.~\ref{fig:prompt_change} in Appendix visually demonstrates the improvements made to the prompt after optimization, illustrating the impact of these refinements.

\section{Experiments }
\subsection{Experimental Setup}  
\textbf{Datasets}: We evaluate SCULPT on four tasks from the Big-Bench Hard (BBH) benchmark \cite{suzgun-etal-2023-challenging}, designed to test LLMs on challenging problems. The selected tasks include \textit{Causal Judgement} (\textbf{CJ}), assessing causal reasoning and moral judgment; \textit{Disambiguation QA} (\textbf{Dis}), resolving ambiguous pronouns; \textit{Formal Fallacies} (\textbf{FF}), distinguishing between valid and fallacious arguments; and \textit{Salient Translation Error Detection} (\textbf{ST}), identifying critical translation errors. Additionally, we evaluate SCULPT on four real-world RAI tasks: \textit{Inappropriate Content Detection} (\textbf{Inapp}), \textit{Hate-Speech Detection} (\textbf{Hate}), \textit{Misinformation Detection} (\textbf{Misinfo}), and \textit{Suicidal Ideation and Drug Use Detection} (\textbf{Selfharm}), each categorized into four harm levels: \textit{No Harm}, \textit{Low Harm}, \textit{Moderate Harm}, and \textit{High Harm}. We also include two multi-label classification tasks with more than ten classes: \textit{GoEmotions} (\textbf{GoE}) \cite{demszky2020goemotions}, classifying Reddit comments into 28 emotion categories, and \textit{BeaverTails} (\textbf{BTails}) \cite{ji2023beavertails}, where human-labeled QA pairs are assigned to multiple categories across 14 harm types.  Table \ref{tab:word_count} in the appendix provides the word counts of the initial prompts, highlighting their length and complexity.

\textbf{Baseline Methods}: We evaluate SCULPT against seven baseline methods. 
(1) \textit{Initial Prompt}, which act as the initial prompt in each optimization method. These prompts for RAI tasks are expert curated (Appendix \ref{sec:initial_prompts}), while those for BBH and multi-label tasks are generated using task descriptions from README files (Appendix \ref{sec:bbh_prompt_gen}). (2) \textit{APE} \cite{zhou2022large}, which generates new prompt candidates by leveraging few-shot examples, then rephrases them to create multiple variations, selecting the best based on validation performance. (3) \textit{LAPE}, a variant of APE, which focuses on generating more descriptive prompts using a predefined template (cf. Appendix \ref{sec:long_ape}). (4) \textit{APEX} \cite{hsieh2023automatic}, which refines prompts by performing sentence-level rephrasing through LLMs while utilizing historical changes for refinement. (5) \textit{OPRO} \cite{yang2023large}, which generates new prompts by relying on historical prompt data and their validation scores. (6) \textit{\protegi} \cite{pryzant2023automatic}, which detects errors in prompts, generates feedback based on these errors, and rephrases the prompts to produce optimized versions.

\textbf{SCULPT Variants:} To assess the impact of reflection aggregation and search space expansion within SCULPT, we evaluate five key variants:  (1) \textit{SCULPT}, which employs \textit{Node-based Aggregation} as the primary method.  (2) \textit{$\text{SCULPT}_{\text{PA}}$}, which replaces \textit{Node-based Aggregation} with \textit{Pattern-based Aggregation}.  (3) \textit{$\text{SCULPT}_{\text{NoAgg}}$}, which omits aggregation entirely to measure the effect of unaggregated reflections on prompt optimization.  (4) Since baselines typically expand the search space through rephrasing, \textit{SCULPT+RP} integrates rephrasing alongside \textit{Node-based Aggregation} to assess the influence of rephrased candidates on SCULPT’s performance. We provide detailed information on rephrasing in Appendix \ref{sec:rephrasing_prompt}. (5) \textit{$\text{SCULPT}_{\text{LAPE}}$}, where the initial prompt is generated using the LAPE method before undergoing optimization in SCULPT. This variant evaluates SCULPT’s performance when it is not initialized with a human-written prompt.

\textbf{Implementation Details}: For most tasks, macro F1 scores are used due to the multiclass nature, while accuracy is reported for \textit{ST}, \textit{Dis} and multi-label tasks. The results reflect the average performance of the top four generated prompts, evaluated across three trials to ensure consistency. All generations were done using \textit{GPT-4o} \cite{openai_gpt4o_2024} with a temperature of 0.5, while the evaluation was performed using both \textit{GPT-4o} and \textit{Llama-3.1-8B} \cite{dubey2024llama}, with a temperature of 0 to guarantee deterministic outputs.

We ensured fairness by assigning all methods the same search budget of 384 total prompt candidates. APE and LAPE generated this number directly, while APEX and OPRO, which produce one prompt per step, were run for 384 steps. \protegi{} was run for 6 steps, producing 64 candidates per step. SCULPT, generating up to 16 candidates per step, could have run for 24 steps, but experiments indicated that performance peaked at 8 steps, with additional steps leading to overfitting. Thus, the reported SCULPT performance reflects the results after 8 optimization steps.

\begin{table}[ht]
\centering
\scriptsize
\setlength\tabcolsep{3pt}
\begin{tabular}{lcccccccccc}
\toprule
\textbf{Method} & \textbf{Dis} & \textbf{CJ} & \textbf{Misinfo} & \textbf{Selfharm} & \textbf{Avg} \\
\midrule
Initial Prompt & 57.3 ± 1.8 & 61.8 ± 1.2 & 36.1 ± 0.8 & 34.9 ± 0.8 & 47.5 \\
APE & 64.3 ± 0.7 & 56.0 ± 1.6 & 33.9 ± 2.3 & 29.4 ± 1.0 & 45.9 \\
LAPE & 60.3 ± 1.9 & 60.0 ± 2.4 & 39.3 ± 2.9 & 33.9 ± 3.2 & 48.4 \\
APEX & 61.5 ± 3.6 & 59.4 ± 3.6 & 28.8 ± 5.0 & 39.6 ± 1.0 & 47.3 \\
OPRO & 49.1 ± 18.4 & 62.9 ± 2.7 & \textbf{43.1 ± 8.3} & 51.8 ± 2.4 & 51.7 \\
ProTeGi & 61.3 ± 3.5 & 58.4 ± 4.3 & 34.5 ± 4.7 & 41.6 ± 1.2 & 49.0 \\
SCULPT & \textbf{65.3 ± 4.3} & \textbf{64.9 ± 1.5} & 37.3 ± 2.8 & \textbf{54.5 ± 5.1} & \textbf{55.5} \\
\bottomrule
\end{tabular}
    \vspace{-1em}
\caption{Performance Comparison using Llama 3.1}
\label{tab:llama3_result}
\end{table}

\begin{figure*}[!ht]
    \centering
    \begin{subfigure}[b]{0.32\textwidth}
        \centering
        \includegraphics[scale=0.15]{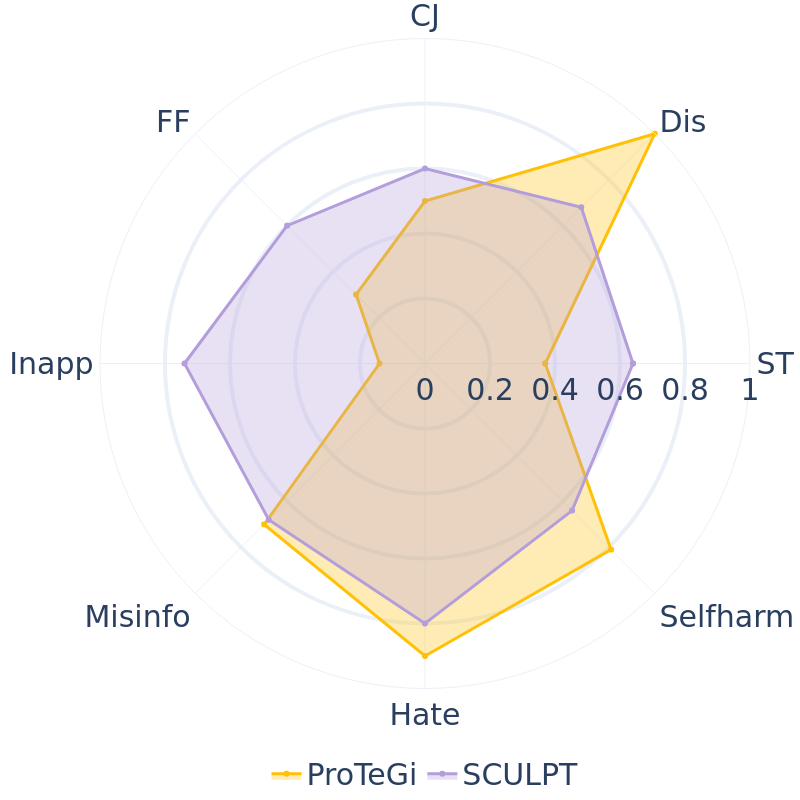}
        \vspace{-0.25em}
        \caption{Textual Similarity}
        \label{fig:radar_texual}
    \end{subfigure}
    \hfill
    \begin{subfigure}[b]{0.32\textwidth}
        \centering
        \includegraphics[scale=0.15]{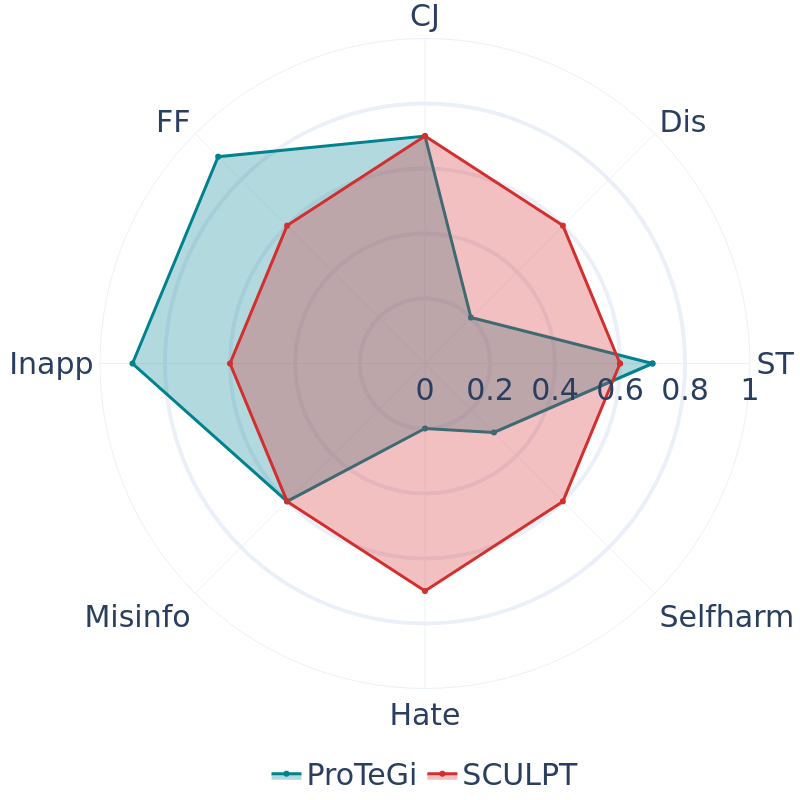}
        \vspace{-0.25em}
        \caption{Semiotic Dissimilarity}
        \label{fig:radar_semiotic}
    \end{subfigure}
    \hfill
    \begin{subfigure}[b]{0.32\textwidth}
        \centering
        \includegraphics[scale=0.15]{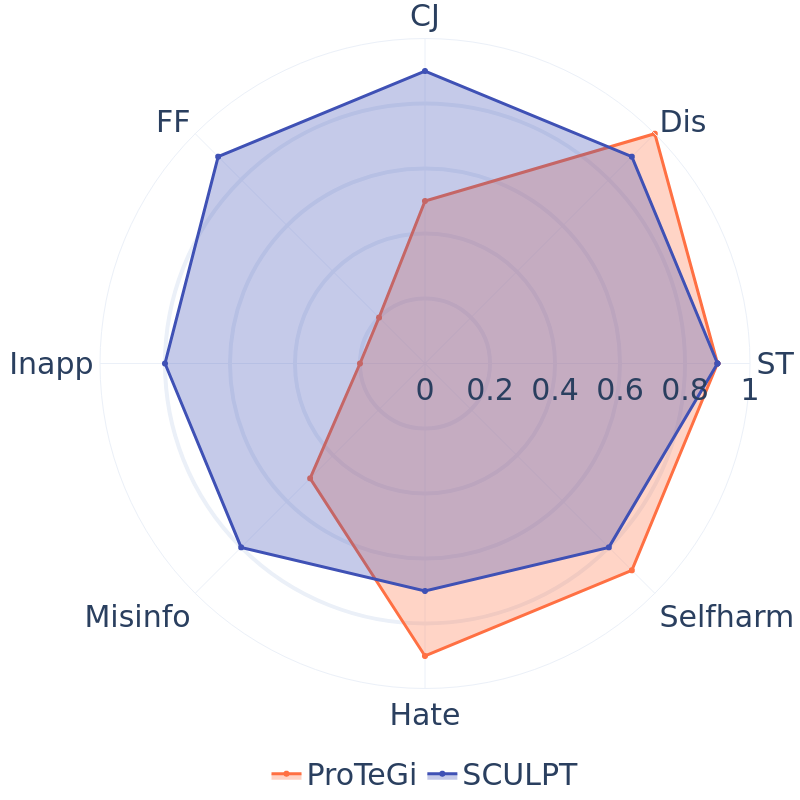}
        \vspace{-0.25em}
        \caption{Information Preservation}
        \label{fig:radar_info_preserve}
    \end{subfigure}
    \vspace{-0.5em}
    \caption{Comparison of Textual Similarity, Semiotic Dissimilarity, and Information Preservation for ProTeGi and SCULPT. ProTeGi exhibits high variability, often making drastic changes while SCULPT maintains stability.}
    \vspace{-1em}
\end{figure*}

\section{Results and Analysis}
\textbf{Performance Comparison with Baselines:} Table \ref{tab:gpt_result} presents the results for {SCULPT} variants and baseline methods across ten tasks using \textit{GPT-4o}. {SCULPT} consistently outperforms all baselines, demonstrating significant improvements over the initial prompt. While APEX struggles to generate meaningful gains, often performing similarly to the initial prompt, OPRO and \protegi{} show minor improvements but lack consistency across different tasks. LAPE performs well on tasks such as \textit{Dis}, \textit{FF} and multi-label tasks surpassing APE and the initial prompt, yet it underperforms in other tasks. On multi-label tasks \textit{GoE} and \textit{BTails}, where a large number of classes makes prompt optimization challenging, most baselines fail to provide substantial improvements. In contrast, {SCULPT} achieves notable performance gains exceeding 10\% on both tasks. Due to space constraints, we have omitted standard deviations here; however, Appendix \ref{sec:full_result} includes them, demonstrating that {SCULPT} exhibits lower variance than other methods, indicating greater stability and reliability across multiple runs. 

Table \ref{tab:llama3_result} presents results for four tasks using \textit{Llama 3.1}. Apart from \textit{Misinfo}, {SCULPT} significantly outperforms all baselines. Interestingly, OPRO delivers better improvements with \textit{Llama-3.1-8B} than with \textit{GPT-4o}, even surpassing {ProTeGi} on 3 out of 4 tasks, suggesting that model-specific behavior influences the effectiveness of optimization strategies. \textit{SCULPT} continues to demonstrate robust performance, reinforcing its adaptability across different models and tasks.

\textbf{Ablation Study of SCULPT}: Table \ref{tab:gpt_result} highlights the performance of different SCULPT variants. {SCULPT}, which uses \textit{Node-based Aggregation}, achieves the best overall results. This variant excels because the Actor can apply all reflections related to a specific node in the prompt simultaneously, ensuring that refinements are comprehensive and targeted. In contrast, \textit{$\text{SCULPT}_{\text{PA}}$} (Pattern-based Aggregation), which clusters reflections based on similarities in erros, may fail to aggregate all reflections for the same node. As a result, some potential improvements for that node may be missed, leading to less precise refinements. While \textit{SCULPT+RP} (rephrasing) delivers results comparable to the standard SCULPT, its impact is inconsistent. Rephrasing does not always lead to further improvements, making it an optional step rather than a core part of the SCULPT.

\begin{table}[ht]
\centering
\scriptsize
\setlength\tabcolsep{3pt}
\begin{tabular}{lcccccccccc}
\toprule
\textbf{Method} & \textbf{SC} & \textbf{OC} & \textbf{OOO} & \textbf{Avg} \\
\midrule
Initial Prompt & 73.1 & 49.0 & 73.3 & 65.13 \\
OPRO & 73.4 & 74.3 & 86.7 & 78.13 \\
ProTeGi & 74.8 & 65.9 & \textbf{100} & 80.23 \\
SCULPT & \textbf{74.9} & \textbf{81.4} & \textbf{100} & \textbf{85.43} \\
\bottomrule
\end{tabular}
    \vspace{-1em}
\caption{Performance Comparison using GPT-4o across short prompt tasks}
\label{tab:shortprompts_perf}
\end{table}

\textbf{Performance Comparison across Short-Prompt Tasks:} To assess the generalizability of {SCULPT}, we evaluate its performance on short-prompt tasks using GPT-4o, a domain typically targeted by algorithms such as OPRO and ProTeGI. As presented in Table \ref{tab:shortprompts_perf}, our evaluation includes tasks from the Big-Bench Hard (BBH) benchmark, \textit{Object Counting} \textbf{(OC)} and \textit{Odd One Out} \textbf{(OOO)}, as well as a generic short-instruction task, \textit{multiclass sentiment classification} \textbf{(SC)}, using Reddit comments from the HuggingFace dataset. Despite being primarily designed for optimizing long prompts with complex structures, SCULPT exhibits strong adaptability, achieving superior performance in \textit{OC} and \textit{SC}, and maintaining competitive results in \textit{OOO}. These findings underscore SCULPT’s robustness and versatility across both complex and simple scenarios.

\begin{table}[ht]
\centering
\scriptsize
\setlength\tabcolsep{4pt}
\begin{tabular}{lcccccccccc}
\toprule
\textbf{Method} & \textbf{Cos Sim} & \textbf{R-1} & \textbf{R-2} & \textbf{R-L} & \textbf{HM} \\
\midrule
Initial Prompt & 76.64 & 37.63 & 14.45 & 23.27 & 35.23 \\
OPRO & 76.28 & 36.16 & 13.28 & 22.08 & 33.86 \\
ProTeGi & 76.84 & 38.73 & 14.59 & 23.85 & 35.92 \\
SCULPT & \textbf{76.96} & \textbf{39.75} & \textbf{15.06} & \textbf{24.59} & \textbf{36.76} \\
\bottomrule
\end{tabular}
    \vspace{-1em}
\caption{Performance Comparison using GPT-4o on a generation task (CNN Sum)}
\label{tab:generative_perf}
\end{table}

\textbf{Performance Comparison across Generative Tasks}: To further evaluate the generalizability of {SCULPT}, we assess its performance on an open-ended generation task in comparison to OPRO and ProTeGi, using GPT-4o as the underlying model. Open-ended tasks typically involve higher ambiguity and demand a nuanced understanding of contextual dependencies to produce high-quality outputs. For this preliminary analysis, we selected the \textit{CNN/DailyMail summarization dataset} \textbf{(CNN Sum)}, where the objective is to generate concise summaries of news articles. Evaluation metrics included \textit{ROUGE-1} \textbf{(R-1)}, \textit{ROUGE-2} \textbf{(R-2)}, \textit{ROUGE-L} \textbf{(R-L)}, \textit{Cosine Similarity} \textbf{(Cos Sim)} computed using a MiniLM sentence transformer, and their \textit{Harmonic Mean} \textbf{(HM)} as an aggregate measure of performance. 

To facilitate more informative feedback, we extended both SCULPT and ProTeGi to pass evaluation metrics such as \textit{R-L} and \textit{Cos Sim} alongside the input, reference summary, and generated output to the Critic module. This enabled the Critic to assess outputs in a more nuanced, non-binary manner. As shown in Table \ref{tab:generative_perf}, SCULPT demonstrates stronger generalization to open-ended generation tasks compared to OPRO and ProTeGi. However, further work is needed to refine Critic's ability to interpret and communicate output quality effectively to the Actor. This remains a critical area for future exploration.

\textbf{Robustness to Prompt Perturbations}: 
To evaluate the robustness of \textsc{SCULPT} under real-world prompt degradations, we design a suite of experiments covering three categories of perturbations: \textbf{semantic}, \textbf{grammatical}, and \textbf{multilingual}. Semantic perturbations include \textit{Localized Perturbations} (LP), where examples are swapped across categories to create subtle inconsistencies, and \textit{Global Perturbations} (GP), which involve swapping entire instruction-example blocks, resulting in severe structural misalignment. Grammatical perturbations are simulated at two levels: \textit{G-Low}, where 20\% of words are misspelled or syntactically degraded, and \textit{G-High}, where the entire prompt is corrupted. Multilingual perturbations involve randomly replacing 20\% of prompt tokens with French (\textit{FR}) or Japanese (\textit{JP}) content to simulate cross-lingual noise. We evaluate these perturbations on two representative classification tasks CJ and Mis, using GPT-4o.

Table~\ref{tab:robustness_combined} summarizes the performance across all perturbation types. \textsc{SCULPT} consistently outperforms both OPRO and ProTeGi across all conditions and both tasks. It demonstrates particular strength in scenarios involving high semantic distortion (GP) and multilingual interference (FR, JP), where other methods degrade sharply. These results reinforce the robustness of \textsc{SCULPT}'s structured refinement mechanism and its practical effectiveness in recovering from both adversarial and naturally occurring prompt degradations.

\begin{table}[h!]
\scriptsize
\centering
\setlength\tabcolsep{4pt}
\begin{tabular}{lccccccc}
\toprule
\textbf{Method} & \textbf{LP} & \textbf{GP} & \textbf{FR} & \textbf{JP} & \textbf{G-Low} & \textbf{G-High} & \textbf{Avg} \\
\midrule
\multicolumn{8}{c}{\textit{Causal Judgment (CJ)}} \\
Initial Prompt   & 71.1 & 71.1 & 71.1 & 71.1 & 71.1 & 71.1 & 71.1 \\
Perturbed Prompt & 69.5 & 70.5 & 67.6 & 69.5 & 69.3 & 69.9 & 69.4 \\
OPRO             & 70.0 & 69.6 & 69.4 & 70.3 & 68.6 & 70.4 & 69.7 \\
ProTeGi          & 68.3 & 69.9 & 70.1 & 73.7 & 70.6 & 71.0 & 70.6 \\
\textbf{SCULPT}  & \textbf{74.1} & \textbf{73.2} & \textbf{75.3} & \textbf{76.7} & \textbf{75.7} & \textbf{73.5} & \textbf{74.7} \\
\midrule
\multicolumn{8}{c}{\textit{Misinformation Detection (Mis)}} \\
Initial Prompt   & 51.5 & 51.5 & 51.5 & 51.5 & 51.5 & 51.5 & 51.5 \\
Perturbed Prompt & 42.2 & 26.5 & 45.8 & 48.4 & 49.6 & 1.5  & 35.6 \\
OPRO             & 48.1 & 43.0 & 47.2 & 37.1 & 50.8 & 45.8 & 45.3 \\
ProTeGi          & 52.4 & 37.8 & 51.8 & 53.3 & 51.5 & 48.5 & 49.2 \\
\textbf{SCULPT}  & \textbf{56.4} & \textbf{50.2} & \textbf{55.9} & \textbf{54.6} & \textbf{57.4} & \textbf{51.6} & \textbf{54.3} \\
\bottomrule
\end{tabular}
\caption{Robustness of SCULPT and baselines under semantic (LP, GP), multilingual (FR, JP), and grammatical (G-Low, G-High) perturbations on CJ and Mis.}
\label{tab:robustness_combined}
\end{table}

\textbf{Optimization using Auto-Generated Prompts}:  In this setting, we evaluate SCULPT’s ability to optimize prompts generated by an automated method rather than a human-crafted prompt. Specifically, we use prompts from LAPE, a structured prompt generation technique, to assess whether SCULPT can refine them to match or surpass expert-designed prompts.  As shown in Table~\ref{tab:gpt_result}, LAPE-generated prompts often perform comparably to or better than human-written ones in BBH and multi-label tasks, where \textit{$\text{SCULPT}_{\text{LAPE}}$} consistently outperforms SCULPT with human-crafted prompts. However, on RAI tasks, where experts carefully designed the initial prompts, \textit{$\text{SCULPT}_{\text{LAPE}}$} does not match SCULPT’s performance but still provides significant improvements over the raw LAPE-generated prompt. These findings highlight SCULPT’s ability to enhance auto-generated prompts, making them a viable alternative when expert-crafted prompts are unavailable.
\begin{figure}[h!]
    \centering
    \includegraphics[scale=0.35]{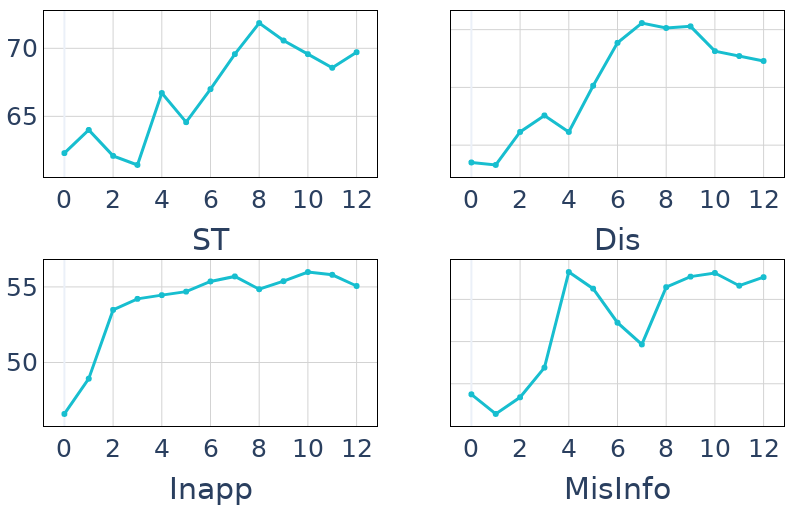}
    \vspace{-1em}
    \caption{Performance across optimization steps}
    \label{fig:roundwise_performance}
    \vspace{-1em}
\end{figure}

\begin{figure*}[!ht]
    \centering
    \begin{subfigure}[b]{0.99\textwidth}
        \centering
        \includegraphics[scale=0.24]{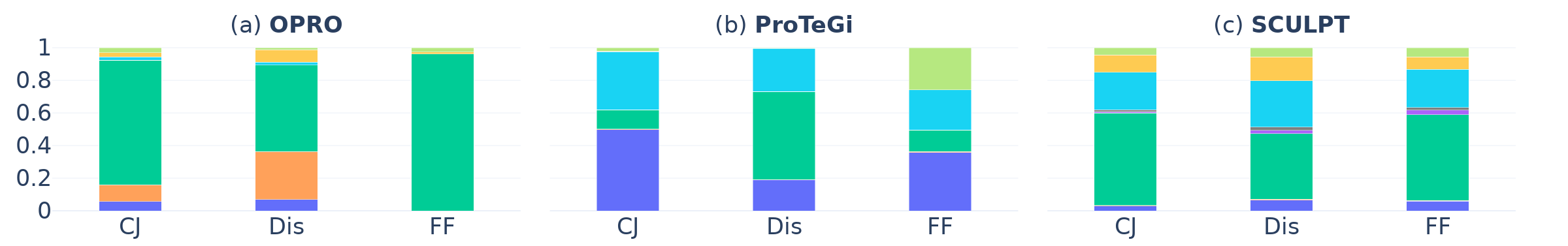}
        \caption{Action distribution across tasks, highlighting SCULPT's consistent application of refinements across different scenarios.}
        \label{fig:action_types_differences}
    \end{subfigure}
    \hfill
    \begin{subfigure}[b]{0.99\textwidth}
        \centering
        \includegraphics[scale=0.14]{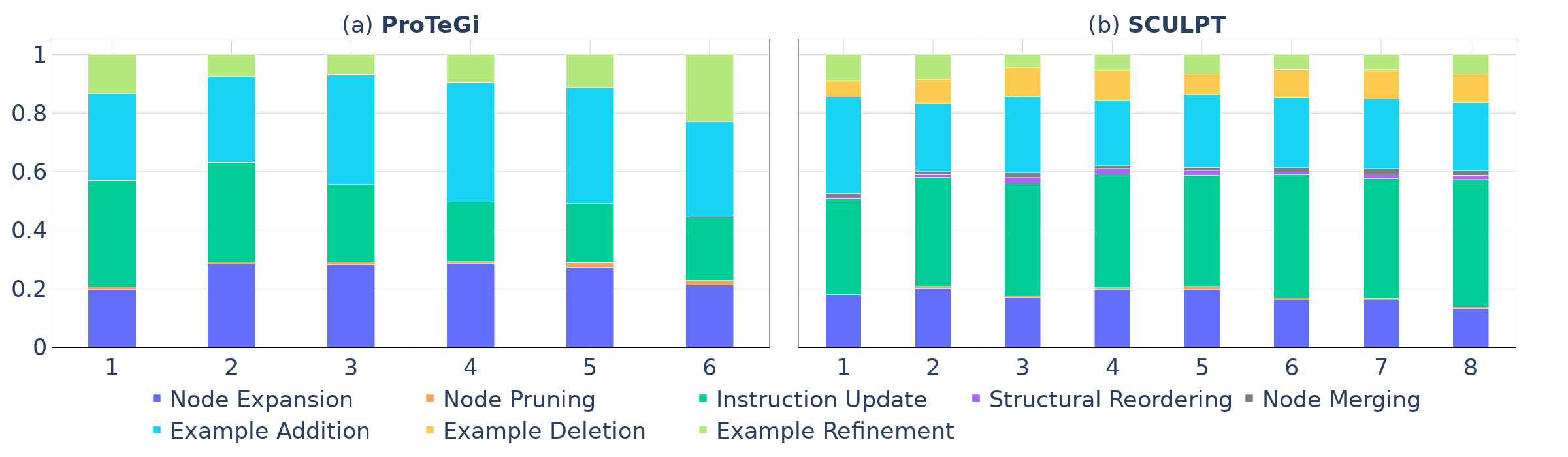}
        \caption{Action distribution over optimization steps, illustrating SCULPT's stable approach to prompt refinement across iterations.}
        \label{fig:roundwise_actions}
    \end{subfigure}
    \vspace{-0.5em}
    \caption{Comparison of action distributions for ProTeGi and SCULPT across tasks (a) and optimization steps (b). SCULPT applies refinements consistently, while ProTeGi exhibits greater variability.}
    \vspace{-1em}
\end{figure*}

\textbf{Performance across Optimization Steps}: To assess the impact of optimization steps, we plot performance after each step in Fig.~\ref{fig:roundwise_performance}. Results show that performance plateaus around step 8 on average. In some cases, continuing beyond 8 steps may lead to overfitting. Based on this, we report performance at the end of 8 steps in our evaluation.

\textbf{Comparative Analysis of Prompts}:  We analyze the structural and semantic differences between initial and optimized prompts using three key metrics. To measure {textual similarity}, we use Sentence Transformers  (\textit{all-MiniLM-L6-v2})  \cite{reimers-gurevych-2019-sentence} to compute semantic overlap. However, due to its 256 tokens input limitation, we create overlapping chunks of the prompts and aggregate their similarity scores to obtain a comprehensive measure. As shown in Fig.~\ref{fig:radar_texual}, SCULPT maintains a similarity score above 0.6 across all tasks, indicating that it applies necessary modifications without drastically altering the original prompt. In contrast, ProTeGi (our best baseline) shows significant variations across tasks, with inconsistent similarity scores, leading to unpredictable prompt modifications.  

\textit{Semiotic dissimilarity} is inversely correlated with textual similarity but provides a more holistic comparison by capturing both semantic and structural differences. Since sentence transformers cannot compare full-length prompts effectively, we employ GPT-4o (cf Appendix \ref{sec:comparative_analysis}) to assess prompt differences at the document level, accounting for logical restructuring, reordering, and coherence beyond surface-level semantic shifts. As shown in Fig.~\ref{fig:radar_semiotic}, ProTeGi exhibits extremely high dissimilarity for \textit{FF} and \textit{Inapp}, reaching values close to 0.9, signifying drastic changes in both content and semantics. This aligns with the observed performance drop from the initial prompt, indicating that excessive modifications can distort task intent. SCULPT, on the other hand, maintains a stable level of dissimilarity across tasks, ensuring that refinements remain controlled and meaningful.

\textit{Information preservation} (Fig.~\ref{fig:radar_info_preserve}) further highlights SCULPT’s consistency in retaining relevant task information. SCULPT systematically removes redundant or misleading content while keeping essential information intact. In contrast, ProTeGi exhibits high variability, occasionally leading to excessive content removal or unintended modifications, which may negatively impact downstream performance. These findings show that SCULPT applies more targeted modifications while ensuring clarity and task relevance.

\textbf{Action Distribution Analysis Across Tasks}: In Figure ~\ref{fig:action_types_differences}, we illustrate the distribution of actions applied by OPRO, ProTeGi, and SCULPT across various tasks. Since OPRO and ProTeGi do not explicitly define their action types, we used LLMs to analyze their behavior and classify changes into predefined action categories (cf. Appendix \ref{sec:opro_protegi_action}). This classification provides a clearer perspective on how these methods refine prompts. SCULPT demonstrates a consistent and balanced distribution of actions across tasks, incorporating \textit{Instruction Updates}, \textit{Example Addition}, \textit{Example Deletion}, and \textit{Node Expansion}. In contrast, OPRO and ProTeGi exhibit significant variability. ProTeGi, for instance, relies heavily on \textit{Node Expansion} ($\sim$40\%) and \textit{Example Addition} ($\sim$30\%), indicating a tendency to resolve prompt issues by adding content, which can lead to overfitting. OPRO, an implicit reflection method, applies less controlled refinements, resulting in more scattered and unsystematic modifications. Similar to our qualitative analysis, we again observe ProTeGi’s inconsistency across tasks, whereas SCULPT consistently applies structured, well-balanced refinements, ensuring stability across diverse tasks.

\textbf{Action Distribution Analysis Across Steps}: Figure~\ref{fig:roundwise_actions} illustrates how action types evolve over optimization steps, averaged across tasks. SCULPT maintains a steady and well-regulated action distribution throughout the steps, ensuring controlled and targeted refinements. In contrast, ProTeGi exhibits high variability, with a growing reliance on \textit{Example Addition} as optimization progresses, potentially leading to overfitting. This evaluation further highlights SCULPT’s stability in contrast to ProTeGi’s inconsistency, reinforcing the trend observed in our qualitative analysis.

\begin{table}[h!]
\scriptsize
\centering
\setlength\tabcolsep{4pt}
\begin{tabular}{llcccccc}
\toprule
\textbf{Task} & \textbf{Method} & \textbf{Input (M)} & \textbf{Comp. (M)} & \textbf{Input \$} & \textbf{Comp. \$} & \textbf{Total \$} \\
\midrule
\multirow{3}{*}{Mis} 
& OPRO     & 1.06 & 0.09 & 2.7 & 0.9 & 3.6 \\
& ProTeGi  & 5.65 & 0.42 & 14.1 & 4.2 & 18.3 \\
& SCULPT   & 0.79 & 0.18 & 2.0 & 1.8 & 3.8 \\
\midrule
\multirow{3}{*}{CJ} 
& OPRO     & 0.36 & 0.04 & 0.9 & 0.4 & 1.3 \\
& ProTeGi  & 2.30 & 0.25 & 5.8 & 2.5 & 8.3 \\
& SCULPT   & 0.64 & 0.18 & 1.6 & 1.8 & 3.4 \\
\bottomrule
\end{tabular}
\caption{Token usage and cost comparison across tasks using GPT-4o. All values are reported in millions (M) of tokens and USD.}
\label{tab:cost_comparison}
\vspace{-1em}
\end{table}

\textbf{Cost Comparison during Optimization}: 
Table~\ref{tab:cost_comparison} compares token usage and cost for \textsc{SCULPT}, ProTeGi, and OPRO on two tasks (Mis, CJ) using GPT-4o. \textsc{SCULPT} is significantly more cost-efficient than ProTeGi while achieving better performance. Compared to OPRO, it has a similar cost in Mis and a slightly higher cost in CJ, due to differences in prompt lengths and method behavior. The OPRO prompt construction strategy, concatenating the top 10 prior prompts, keeps the input size low for short tasks such as CJ but increases the cost for longer tasks like Mis. SCULPT, by iteratively refining a single structured prompt, maintains a stable and efficient cost profile across settings. Although SCULPT and ProTeGi both use iterative feedback, ProTeGi’s unstructured approach leads to over 2$\times$ higher cost. SCULPT also uses more completion tokens than OPRO, but since input tokens dominate overall cost, the impact is minimal. These results confirm SCULPT’s ability to offer strong performance with practical cost-efficiency.

\section{Conclusion}  
We introduce SCULPT, a novel framework for optimizing long prompts in LLMs through hierarchical structuring and targeted refinements. Unlike existing methods which struggle with complex multi-instruction prompts, SCULPT applies structured modifications while maintaining a balanced distribution of actions, ensuring controlled and high-quality refinements. It demonstrates strong robustness against prompt perturbations, outperforming existing methods in handling adversarial modifications.  \textit{SCULPT} effectively refines both expert-curated and auto-generated prompts, achieving strong performance across multiple tasks. Our comparative analysis highlights its ability to preserve key information while systematically improving clarity and coherence. Additionally, \textit{SCULPT} reduces computational costs by 50\% compared to ProTeGi, making it a scalable and resource-efficient approach. These results position \textit{SCULPT} as a reliable solution for enhancing LLM performance across diverse tasks.

\section{Limitations}  
While \textit{SCULPT} demonstrates strong performance, it has certain limitations. Our evaluation is restricted to two LLMs, GPT-4o and Llama 3.1, due to computational constraints. A broader study across diverse LLM sizes and architectures could provide deeper insights into its generalizability and effectiveness at different scales. Additionally, \textit{SCULPT} has only been tested on English-language prompts. Extending it to multilingual settings would enhance its applicability to global contexts and broader tasks. Future work could explore leveraging historical optimization trajectories to guide refinements, enabling \textit{SCULPT} to learn from previous iterations and dynamically adjust modifications based on past improvements. Integrating memory-based or reinforcement learning techniques could enhance adaptability, reducing unnecessary modifications and improving efficiency over multiple optimization cycles.  
\bibliography{references}

\newpage
\appendix

\addcontentsline{toc}{section}{Appendix} 
\part{Appendix} 
\parttoc

\begin{figure*}[!ht]
    \centering
    \includegraphics[scale=0.09]{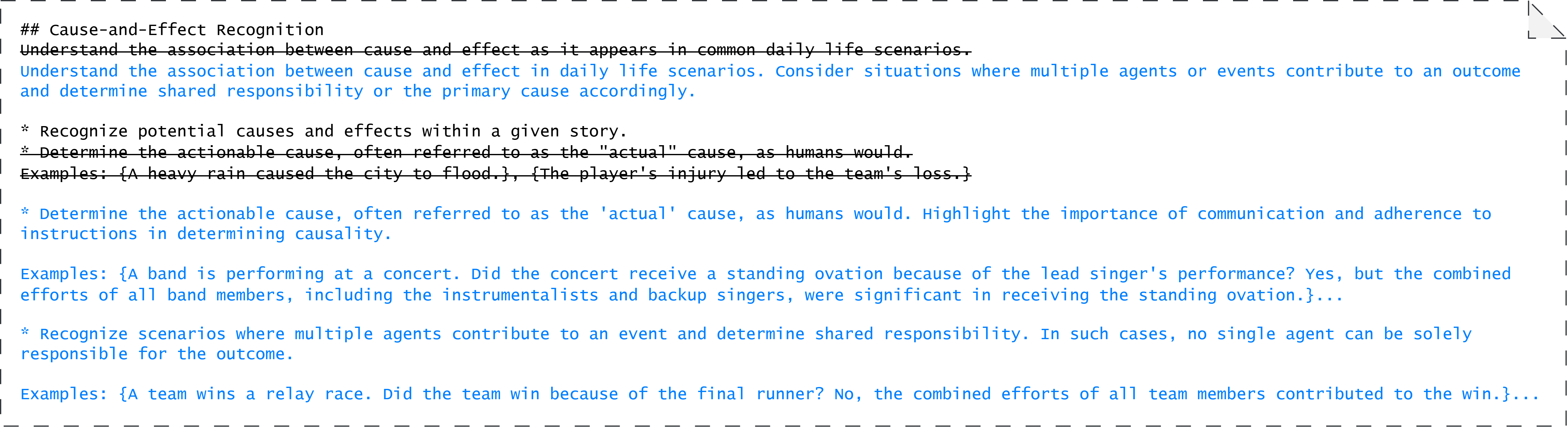}
        \caption{Edits applied to the prompt using SCULPT, where strikethrough represents removed content and blue text indicates additions. These modifications involve example addition, node expansion, and instruction update.}
    \label{fig:prompt_change}
    \vspace{-0.5em}
\end{figure*}

\begin{table}[h!]
\centering
\scriptsize
\setlength\tabcolsep{10pt}
\begin{tabular}{lc}
\toprule
\textbf{Task} & \textbf{\# Words} \\
\midrule
Formal Fallacy  & 382 \\
Causal Judgement  & 367 \\
Salient Translation & 279 \\
Disambiguation & 346 \\
Inappropriate & 2644 \\
Hate & 1554 \\
Misinformation & 1335 \\
SelfHarm & 933 \\
BeaverTails & 366 \\
GoEmotions & 509 \\
\bottomrule
\end{tabular}
\caption{Number of words in the initial prompts}
\label{tab:word_count}
\end{table}

\begin{table}[h!]
\centering
\scriptsize 
\begin{tabular}{lcccc}
\toprule
\textbf{Method} & \textbf{ST} & \textbf{Dis} & \textbf{CJ} & \textbf{FF} \\
\midrule
Initial Prompt  & 62.3 ± 1.0 & 74.1 ± 0.7 & 71.1 ± 0.1 & 80.2 ± 1.4 \\
APE             & 51.0 ± 1.6 & 74.3 ± 1.1 & 71.8 ± 2.6 & 75.3 ± 2.5 \\
LAPE        & 52.7 ± 4.0 & 78.0 ± 1.0 & 72.2 ± 3.3 & 81.3 ± 0.3 \\
APEX            & 62.7 ± 0.6 & 61.5 ± 2.2 & 70.5 ± 0.7 & 80.6 ± 1.3 \\
OPRO            & 64.6 ± 2.0 & 75.1 ± 1.4 & 72.7 ± 1.1 & 80.7 ± 2.1 \\
ProTeGi         & 65.5 ± 3.4 & 74.8 ± 1.2 & 68.9 ± 2.1 & 70.1 ± 3.9 \\
\midrule
$\text{SCULPT}_{\text{NoAgg}}$       & 65.2 ± 1.1 & 77.3 ± 1.2 & 75.4 ± 2.4 & 83.1 ± 1.1 \\
$\text{SCULPT}_{\text{PA}}$       & 66.2 ± 2.1 & 77.6 ± 1.9 & \textbf{76.9 ± 1.9} & 83.7 ± 1.1 \\
SCULPT+RP       & 67.6 ± 1.9 & 78.0 ± 0.6 & 75.1 ± 2.0 & {84.7 ± 1.1} \\
SCULPT          & \textbf{68.8 ± 1.5} & {80.1 ± 1.9} & 75.9 ± 1.5 & 83.7 ± 2.5 \\
$\text{SCULPT}_{\text{LAPE}}$           & {66.8 ± 2.2} & \textbf{81.1 ± 2.4} & 76.1 ± 1.9 & \textbf{86.5 ± 2.7} \\
\bottomrule
\end{tabular}
\vspace{-1em} 
\caption{Performance comparison using GPT-4o on BBH tasks}
\label{tab:gpt_result_bbh}
\end{table}

\begin{table}[h!]
\centering
\scriptsize
\begin{tabular}{lcccc}
\toprule
\textbf{Method} & \textbf{Inapp} & \textbf{Misinfo} & \textbf{Hate} & \textbf{Selfharm} \\
\midrule
Initial Prompt  & 46.6 ± 1.3 & 51.5 ± 0.6 & 46.8 ± 0.1 & 66.4 ± 0.5 \\
APE             & 45.3 ± 0.4 & 31.9 ± 5.9 & 29.3 ± 2.7 & 38.1 ± 0.4 \\
LAPE        & 42.4 ± 0.9 & 35.4 ± 2.1 & 37.6 ± 0.5 & 44.2 ± 1.3 \\
APEX            & 48.0 ± 0.4 & 50.5 ± 0.8 & 47.1 ± 0.3 & 65.2 ± 0.7 \\
OPRO            & 46.6 ± 2.2 & 51.0 ± 4.0 & 40.9 ± 3.6 & 65.1 ± 2.1 \\
ProTeGi         & 44.8 ± 5.4 & 54.8 ± 1.2 & 51.9 ± 1.6 & 66.5 ± 2.1 \\
\midrule
$\text{SCULPT}_{\text{NoAgg}}$       & 53.6 ± 0.7 & 53.6 ± 2.5 & 51.9 ± 0.3 & 65.5 ± 0.9 \\
$\text{SCULPT}_{\text{PA}}$       & \textbf{55.3 ± 0.7} & \textbf{56.7 ± 0.8} & \textbf{53.1 ± 0.3} & 68.8 ± 0.3 \\
SCULPT+RP       & 55.0 ± 0.8 & 55.3 ± 2.4 & 52.9 ± 1.4 & \textbf{69.0 ± 1.8} \\
SCULPT          & 55.0 ± 0.8 & 54.9 ± 0.6 & \textbf{53.1 ± 1.2} & 68.5 ± 1.0 \\
$\text{SCULPT}_{\text{LAPE}}$           & 48.2 ± 1.5 & 48.8 ± 2.3 & 44.5 ± 3.1 & 61.8 ± 1.4 \\
\bottomrule
\end{tabular}
\caption{Performance comparison using GPT-4o on RAI tasks}
\label{tab:gpt_result_rai}
\end{table}

\begin{table}[h!]
\centering
\scriptsize
\begin{tabular}{lcc}
\toprule
\textbf{Method} & \textbf{BTails} & \textbf{GoE} \\
\midrule
Initial Prompt  & 41.8 ± 0.3 & 7.8 ± 0.4 \\
APE             & 46.4 ± 3.4 & 0 ± 0\\
LAPE        & 39.1 ± 2.1 & 19.6 ± 1.3 \\
APEX            & 40.7 ± 1.8 & 8.0 ± 0.1 \\
OPRO            & 41.5 ± 1.4 & 11.8 ± 2.1 \\
ProTeGi         & 45.3 ± 0.9 & 17.8 ± 0.1 \\
\midrule
$\text{SCULPT}_{\text{NoAgg}}$       & 49.6 ± 0.5 & 29.8 ± 1.9 \\
$\text{SCULPT}_{\text{PA}}$       & 49.3 ± 0.6 & 29.6 ± 1.3 \\
SCULPT+RP       & 49.6 ± 0.6 & 22.4 ± 2.1 \\
SCULPT          & 50.5 ± 0.4 & 30.6 ± 0.5 \\
$\text{SCULPT}_{\text{LAPE}}$           & 50.5 ± 2.6 & 30.6 ± 1.1 \\
\bottomrule
\end{tabular}
\caption{Performance comparison using GPT-4o on GoEmotions and BeaverTails}
\label{tab:gpt_result_multilabel}
\end{table}

\section{Detailed Results with Standard Deviation}
\label{sec:full_result}
In Tables \ref{tab:gpt_result_bbh}, \ref{tab:gpt_result_rai} and \ref{tab:gpt_result_multilabel}, we have reported the mean and standard deviation of the performances of every method across three runs using GPT-4o. From these tables, it is evident SCULPT provides least variance across runs compared to other methods. Automatic prompt generative approaches APE and LAPE provide higher variance compared to prompt optimization methods.

\section{UCB-based Prompt Selection Strategy}
\label{sec:ucb}
Evaluating the generated candidate prompts on the validation set $\mathcal{D}_{\text{val}}$ is a computationally expensive process. To minimize these computations, we have used the Upper Confidence Bound (UCB) Bandit algorithm as mentioned in \cite{pryzant2023automatic}. This helps to minimize the number of prompts to be evaluated as well as the number of validation set samples to evaluate them on. This is done based on the proposal distribution of prompt performance which is updated after each evaluation round. At the end top $b$ prompts with highest weight in the distribution are selected.

See Algorithm \ref{algo:ucb} for details, where $Q_t(p_i)$ is the estimated performance of prompt $p_i$ at time step $t$, $N_t(p_i)$ is the total queries for prompt $i$ so far at time $t$, and $c$ is the exploration parameter.

\begin{algorithm}[h!]
\scriptsize
\caption{UCB Bandits Candidate Selection}
\begin{algorithmic}
\State \textbf{Require} $n$ prompts $p_1, p_2, ..., p_n$, dataset $\mathcal{D}_{\text{val}}$, $T$ time steps and metric function $m$
\State Initialize: $N_t(p_i) \gets 0$ for all $i$ = 1,...,$n$
\State Initialize: $Q_t(p_i) \gets 0$ for all $i$ = 1,...,$n$
\For{$t = 1,...,T$}
    \State Sample uniformly $\mathcal{D}_{\text{sample}} \subset \mathcal{D}_{\text{val}}$
    \State $p_i \gets \argmax{_p Q_t(p) + c \sqrt{\frac{\log t}{N_t(p)}}}$
    \State Observe reward $r_i,_t = m(p_i, \mathcal{D}_{\text{sample}})$
    \State $N_t(p_i) \gets N_t(p_i) + |\mathcal{D}_{\text{sample}}|$
    \State $Q_t(p_i) \gets Q_t(p_i) + \frac{r_i,_t}{N_t(p_i)}$
\EndFor
\State \textbf{return} $SelectTop_b(Q_T)$
\end{algorithmic}
\label{algo:ucb}
\end{algorithm}

\section{Task and Initial Prompt Statistics}
Table \ref{tab:data_info} presents the number of examples in the training, validation, and test sets for each task, offering an overview of the dataset size. Additionally, Table \ref{tab:word_count} in Appendix lists the word counts of the initial prompts used in each task, highlighting the length and complexity of these prompts. This information emphasizes the challenges posed by long and unstructured prompts, which require systematic optimization to ensure model performance. We have provided the list of initial prompts in Section \ref{sec:initial_prompts}.

\begin{table}[!t]
    \centering
    \small
    \begin{tabular}{lccc}
    \toprule
       Dataset & Validation & Train & Test \\ \midrule
       Causal Judgement  & 19 & 36 & 129 \\
       Disambiguation QA  & 24 & 49 & 174 \\
       Formal Fallacies  & 24 & 49 & 174 \\
       Salient Translation  & 24 & 49 & 174 \\
       Inappropriate  & 122 & 242 & 851 \\
       Misinformation  & 122 & 242 & 851 \\
       Hate & 122 & 242 & 851 \\
       SelfHarm   & 122 & 242 & 851 \\
       BeaverTails   & 1020 & 1000 & 1000 \\
       GoEmotions   & 3426 & 1000 & 1000 \\       
    \bottomrule
    \end{tabular}
    \caption{Dataset size information}
    \label{tab:data_info}
\end{table}

\begin{figure*}[!htbp]
\begin{subfigure}{0.99\textwidth}
  \centering   \captionsetup{justification=centering} 
  \includegraphics[scale=0.2]{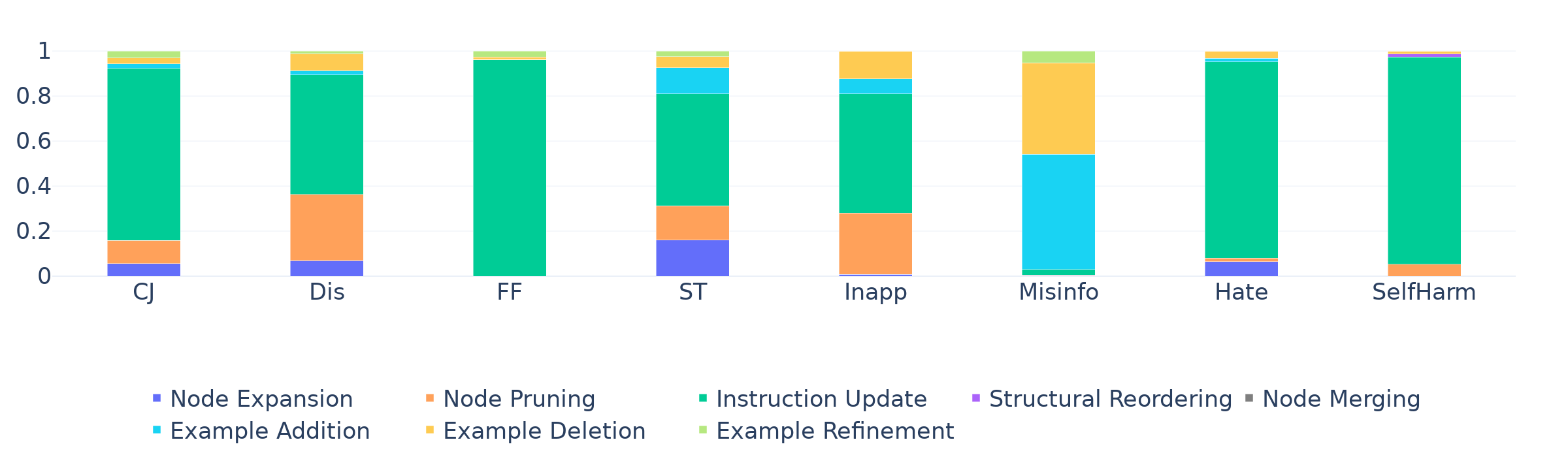}
  \caption{OPRO}
\end{subfigure}
\hfill
\begin{subfigure}{0.99\textwidth}
  \centering   \captionsetup{justification=centering} 
  \includegraphics[scale=0.2]{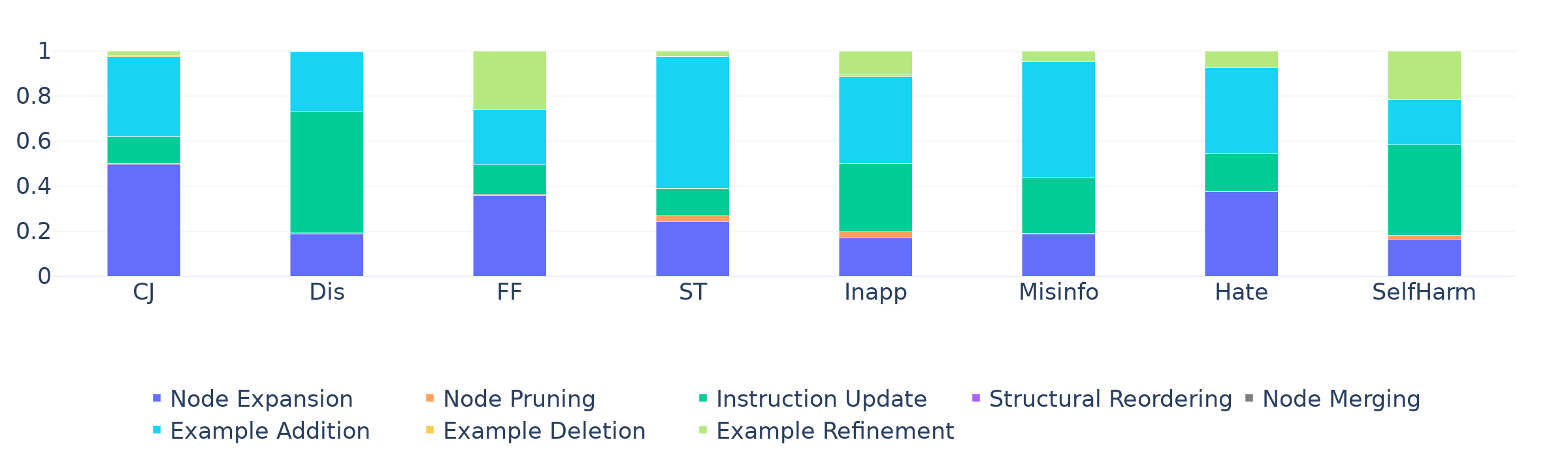}
  \caption{\protegi}
\end{subfigure}
\hfill
\begin{subfigure}{0.99\textwidth}
  \centering   \captionsetup{justification=centering} 
  \includegraphics[scale=0.2]{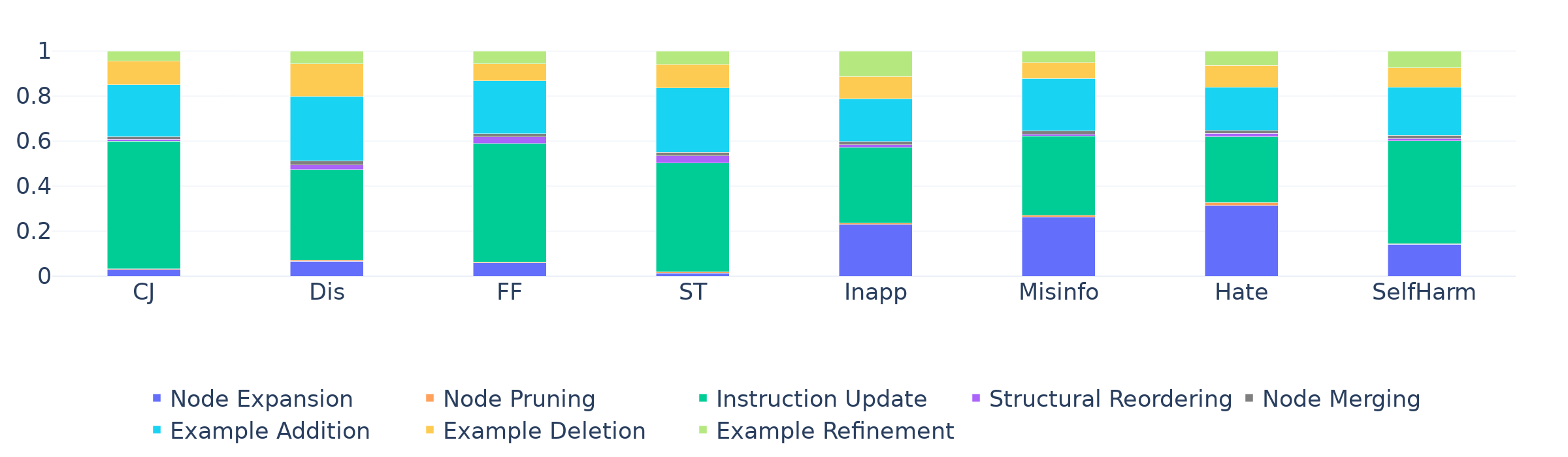}
  \caption{SCULPT}
\end{subfigure}
\caption{Action distribution across tasks of OPRO, ProTeGi and SCULPT.}
\label{fig:taskwise_actions}
\end{figure*}

\section{Critic and Actor Interactions in SCULPT}
\label{sec:critic_actor_response}

This section illustrates the interactions between the Critic and Actor modules within SCULPT by presenting both the preliminary and error-assessment reflections, the Actor's responses to each type of feedback, and the resulting prompt updates. Specifically, we showcase how these actions contribute to prompt refinements during the first round of the Salient Translation task, demonstrating the iterative role of both modules in improving the prompt's clarity and alignment with task requirements.
\vspace{1em}

\hspace{-1em}\textbf{Critic's Preliminary Assessment}: Table \ref{tab:prelim-critic-reponse-ST} shows the preliminary feedback provided in round 1. The feedback identifies multiple areas for improvement, including adding examples for different types of translation errors and rephrasing certain parts of the prompt to enhance clarity and relevance.
\vspace{1em}

\hspace{-1em}\textbf{Actor Response to Preliminary Assessment}: The Actor module processes the Critic's feedback and suggests a set of actions, which are summarized in Table \ref{tab:prelim-actor-reponse-ST}. One key action involves adding specific examples for ``Named Entities'' errors, while another focuses on rephrasing the task description in the `Task > body' section for greater clarity.
\vspace{1em}

\hspace{-1em}\textbf{Updated Prompt Based on Preliminary Assessment}: After applying the Preliminary Assessment, notable improvements are observed in the prompt. The task description has been rephrased for clarity, and new examples for each translation error type have been added. See Table \ref{tab:prompt-after-prelim} for the updated prompt, which shows significant refinements compared to the initial prompt in Table \ref{prompt: salient-long}.
\vspace{1em}

\hspace{-1em}\textbf{Critic's Error Assessment}: Table \ref{tab:error-critic-reponse-ST} provides the error assessment based on the Critic's evaluation of the prompt in response to specific model errors. The reflection highlights areas where new examples need to be added and suggests rephrasing sections to clarify definitions of error types, ensuring fewer ambiguities.
\vspace{1em}

\hspace{-1em}\textbf{Actor Response to Error Assessment}: In response to the error-assessment feedback, the Actor module suggests targeted actions, which are listed in Table \ref{tab:error-assess-actor-reponse-ST}. These include adding examples for error type 1 and rephrasing sections as needed to avoid confusion and improve clarity.
\vspace{1em}

\hspace{-1em}\textbf{Updated Prompt Based on Error Assessment}: The updated prompt, following both initial assessments and error analysis from the first round, is presented in Table  \ref{tab:prompt-after-error-assess}. In contrast to the original version (Table \ref{prompt: salient-long}), the revised prompt integrates additional examples and restructured sections. This demonstrates the capability of \SCULPT to systematically refine prompts through controlled reflections and targeted adjustments.

\vspace{1em}

\hspace{-1em}\textbf{Final Prompt Post-Optimization}: The fully optimized prompt, after the entire SCULPT optimization process, is presented in Table \ref{tab:prompt-after-process-completion}. This refined version shows significant improvements over the initial prompt (Table \ref{prompt: salient-long}). Key enhancements include a clearer redefinition of error categories, refined examples, and improved clarity in the final output classes. These adjustments, made after the Preliminary and Error Assessment updates (refer to Table \ref{tab:prompt-after-error-assess}), ensure that the user is guided more effectively through the task. This final prompt underscores the systematic nature of SCULPT’s iterative refinement process, showcasing its ability to optimize prompt design efficiently and effectively.

\section{Action Types Distribution across tasks and steps}
\label{sec:action_types}
In Fig.~\ref{fig:taskwise_actions}, we present the distribution of action types applied by OPRO, ProTeGi, and SCULPT across different tasks, illustrating how each method handles prompt refinements.

\begin{figure*}[!htbp]
\begin{subfigure}{0.99\textwidth}
\centering   \captionsetup{justification=centering} 
  \includegraphics[scale=0.2]{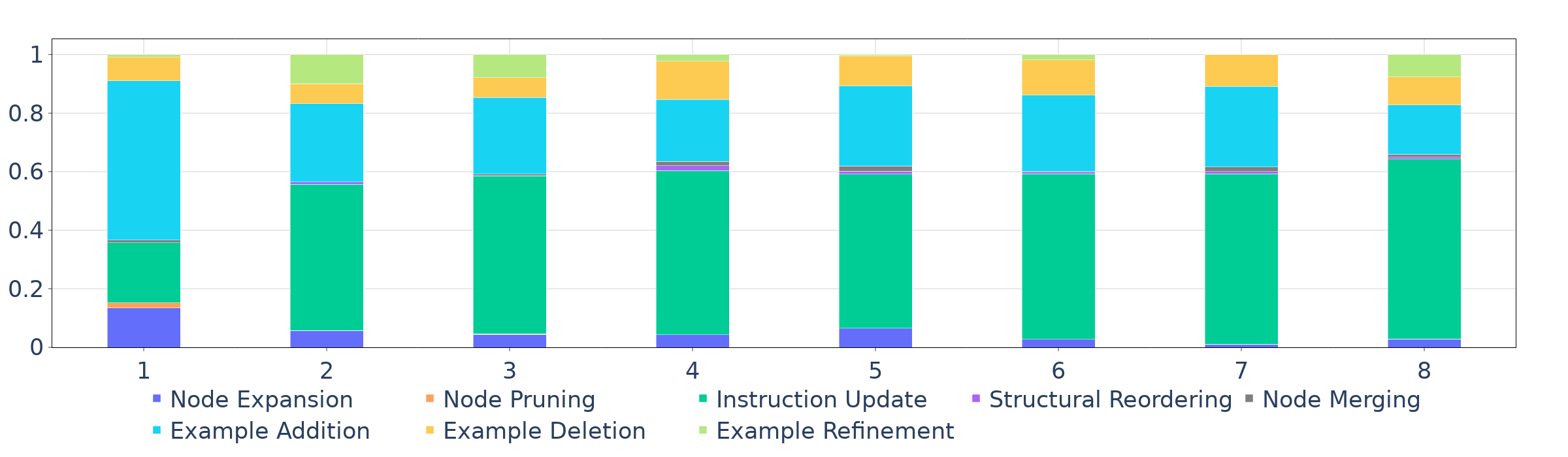}
  \caption{Causal Judgement}
\end{subfigure}
\hfill
\begin{subfigure}{0.99\textwidth}
  \centering   \captionsetup{justification=centering} 
  \includegraphics[scale=0.2]{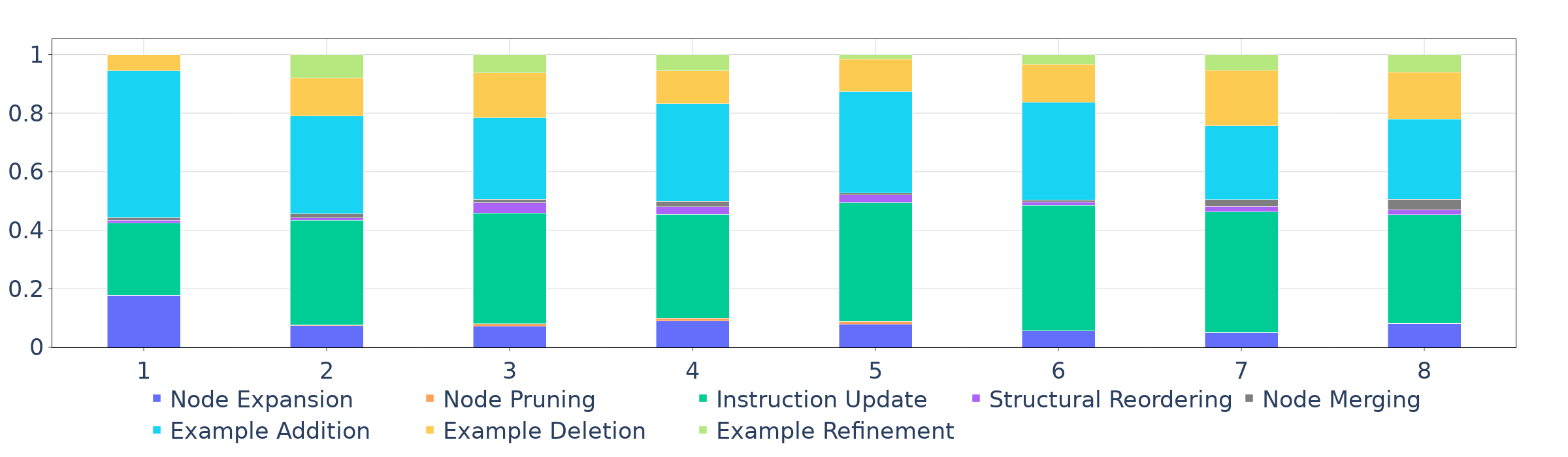}
  \caption{Disambiguation}
\end{subfigure}
\hfill
\begin{subfigure}{0.99\textwidth}
  \centering   \captionsetup{justification=centering} 
  \includegraphics[scale=0.2]{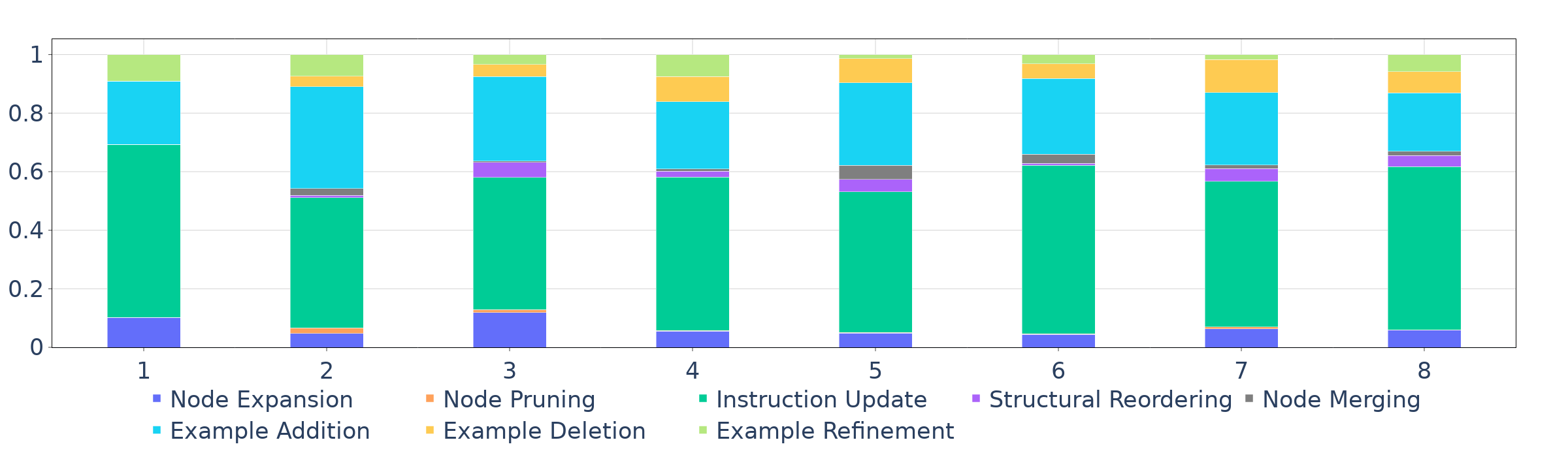}
  \caption{Formal Fallacy}
\end{subfigure}
\hfill
\begin{subfigure}{0.99\textwidth}
  \centering   \captionsetup{justification=centering} 
  \includegraphics[scale=0.2]{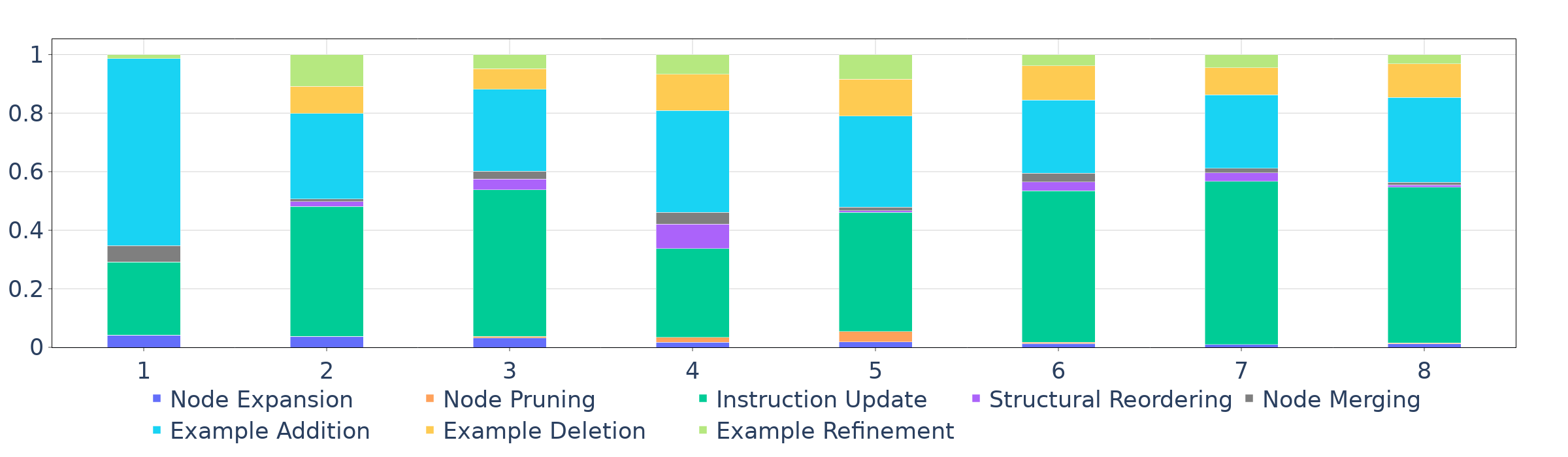}
  \caption{Salient Translation}
\end{subfigure}
\caption{Action distribution over optimization steps in SCULPT}
\label{fig:taskwise_sculpt_1}
\end{figure*}

\begin{figure*}[!htbp]
\begin{subfigure}{0.99\textwidth}
\centering   \captionsetup{justification=centering} 
  \includegraphics[scale=0.2]{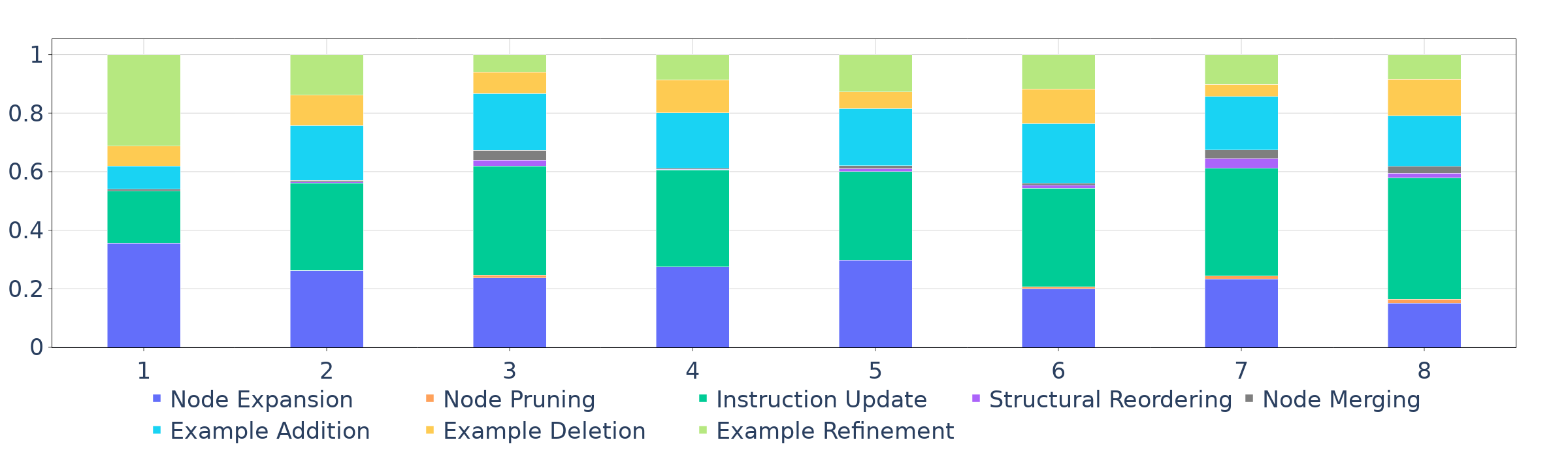}
  \caption{Inappropriate}
\end{subfigure}
\hfill
\begin{subfigure}{0.99\textwidth}
  \centering   \captionsetup{justification=centering} 
  \includegraphics[scale=0.2]{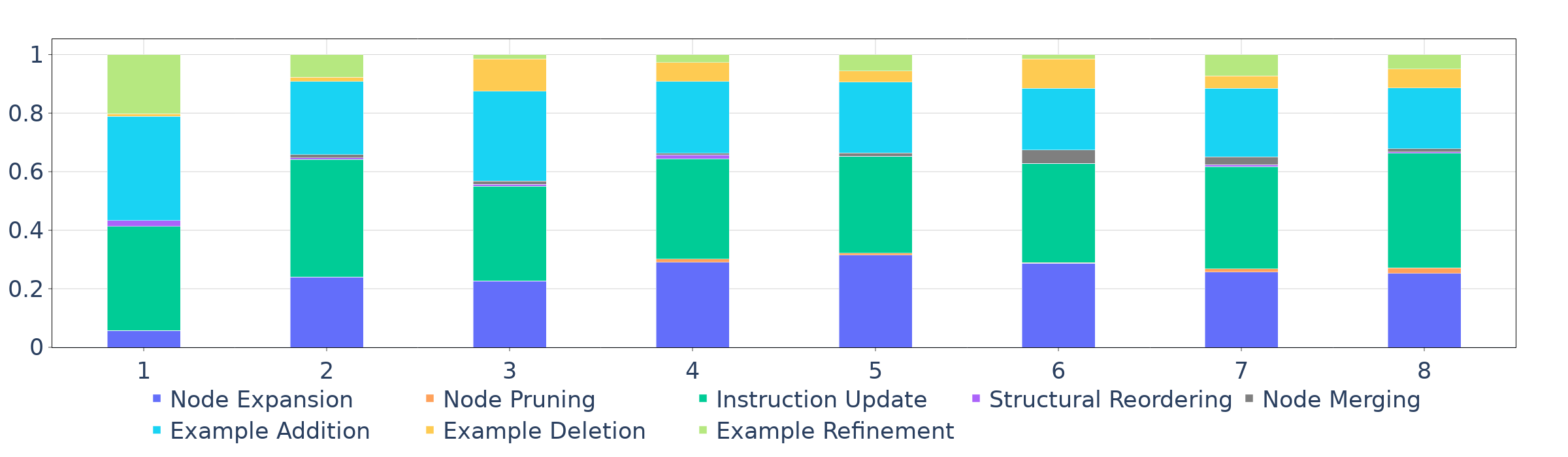}
  \caption{Misinformation}
\end{subfigure}
\hfill
\begin{subfigure}{0.99\textwidth}
  \centering   \captionsetup{justification=centering} 
  \includegraphics[scale=0.2]{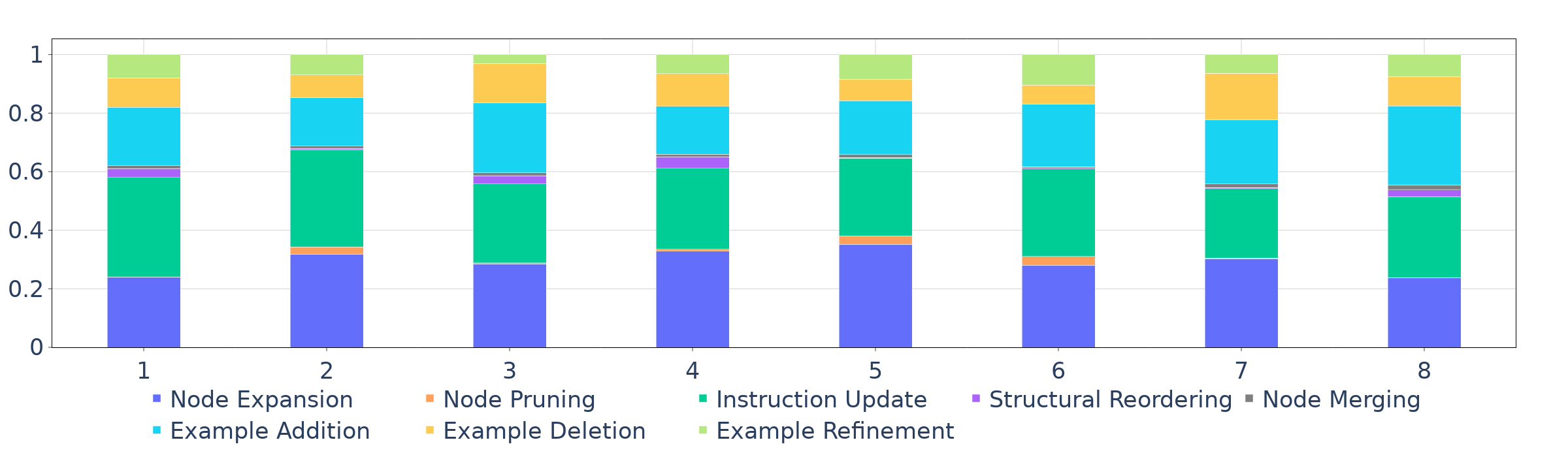}
  \caption{Hate}
\end{subfigure}
\hfill
\begin{subfigure}{0.99\textwidth}
  \centering   \captionsetup{justification=centering} 
  \includegraphics[scale=0.2]{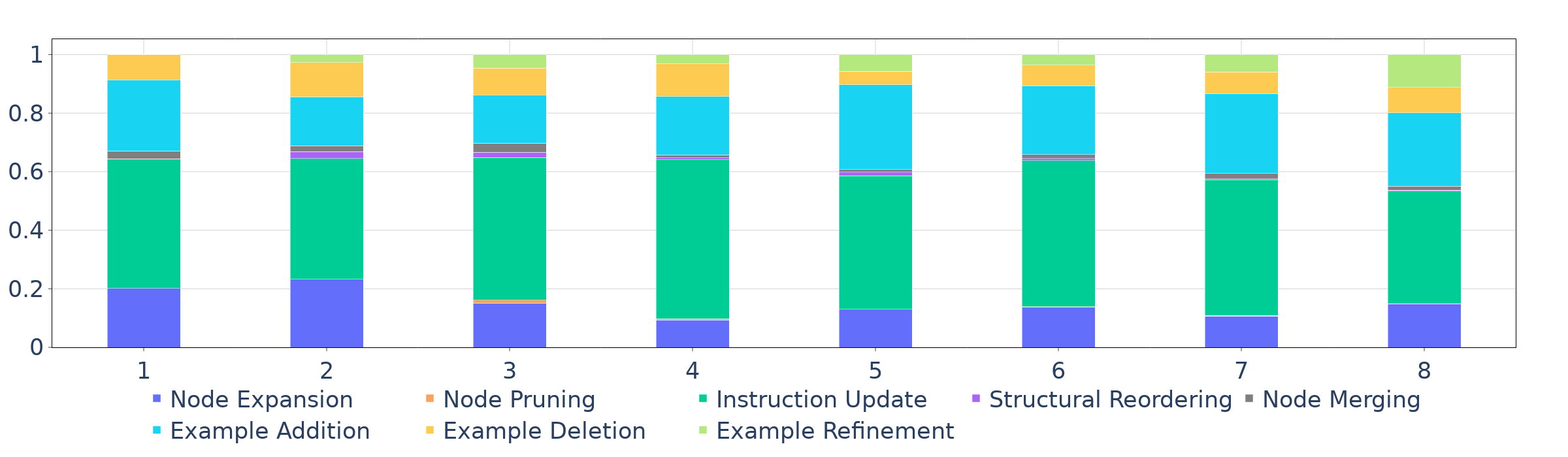}
  \caption{SelfHarm}
\end{subfigure}
\caption{Action distribution over optimization steps in SCULPT}
\label{fig:taskwise_sculpt_2}
\end{figure*}

\begin{figure*}[!htbp]
\begin{subfigure}{0.99\textwidth}
\centering   \captionsetup{justification=centering} 
  \includegraphics[scale=0.2]{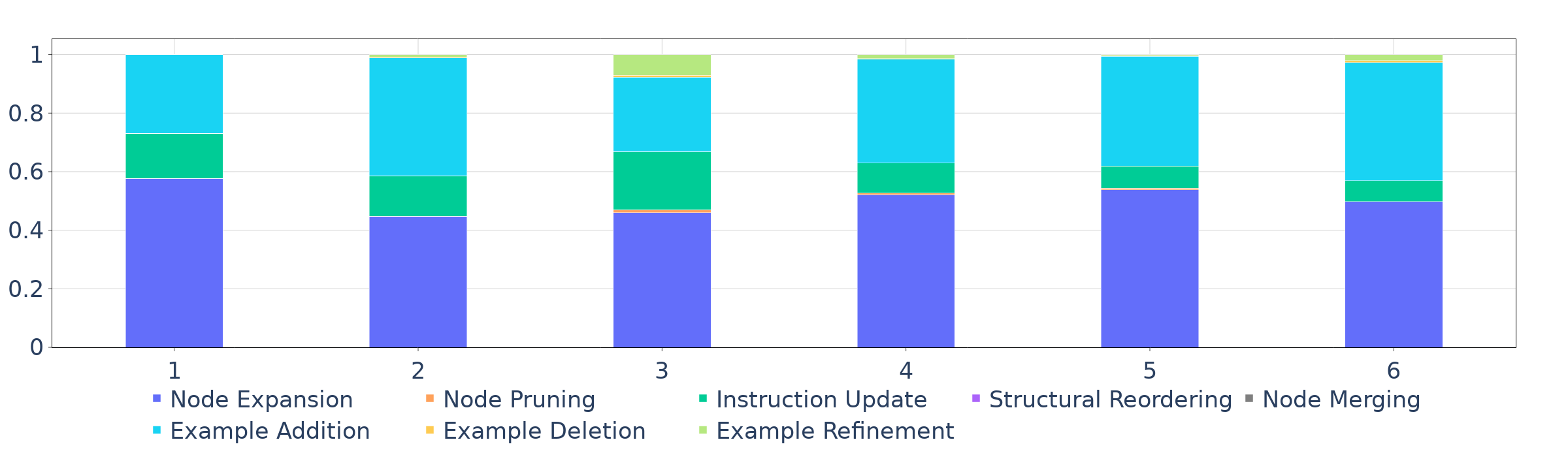}
  \caption{Causal Judgement}
\end{subfigure}
\hfill
\begin{subfigure}{0.99\textwidth}
  \centering   \captionsetup{justification=centering} 
  \includegraphics[scale=0.2]{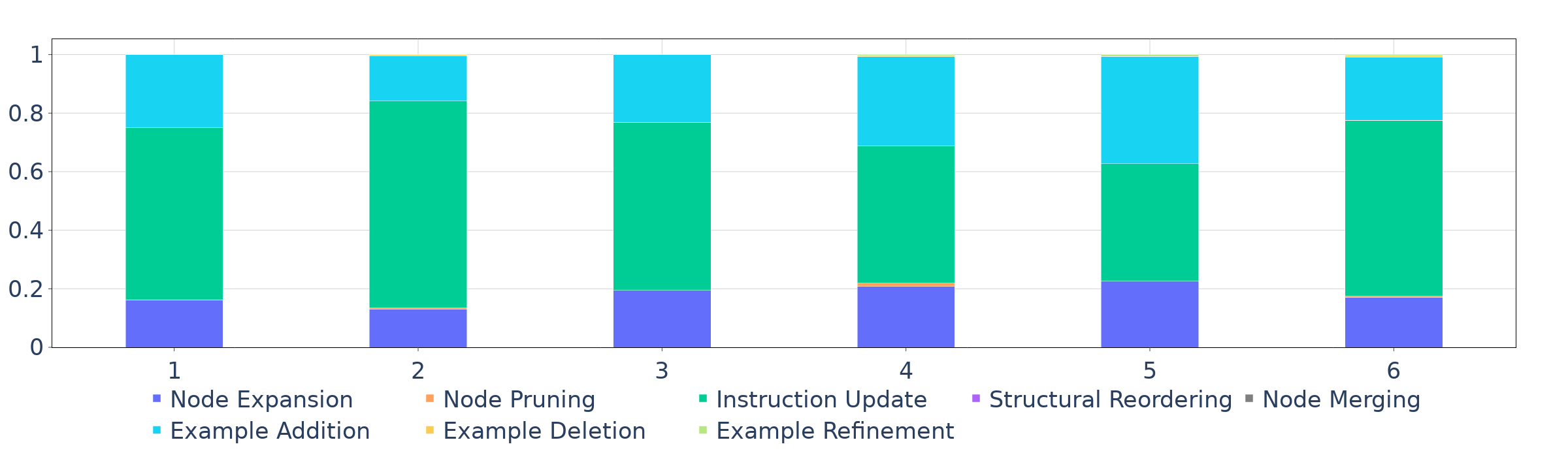}
  \caption{Disambiguation}
\end{subfigure}
\hfill
\begin{subfigure}{0.99\textwidth}
  \centering   \captionsetup{justification=centering} 
  \includegraphics[scale=0.2]{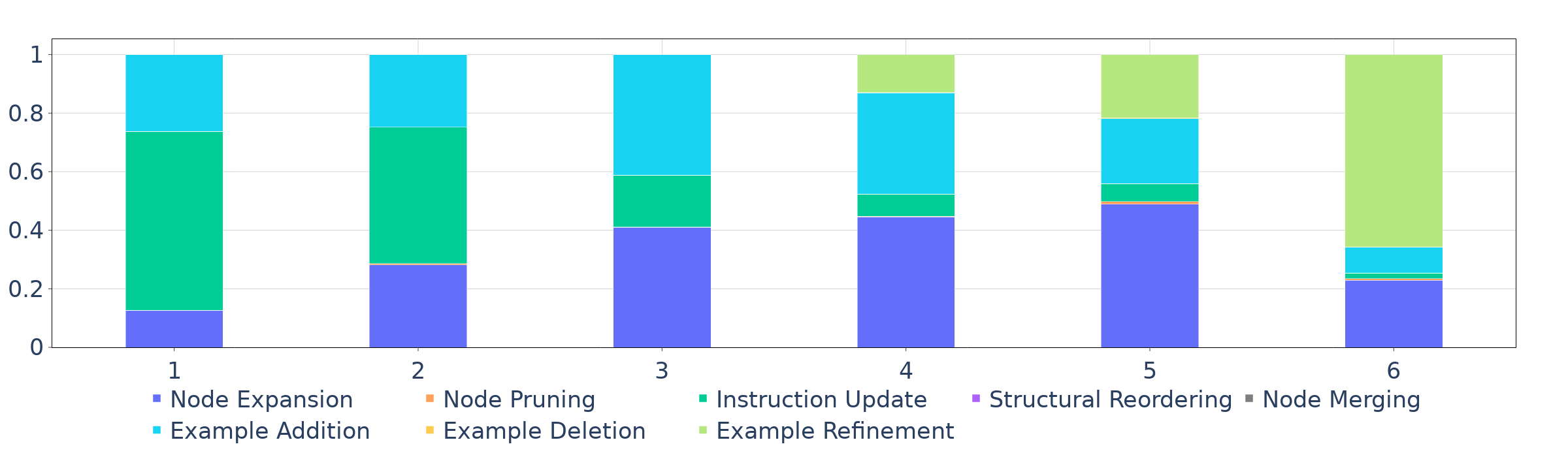}
  \caption{Formal Fallacy}
\end{subfigure}
\hfill
\begin{subfigure}{0.99\textwidth}
  \centering   \captionsetup{justification=centering} 
  \includegraphics[scale=0.2]{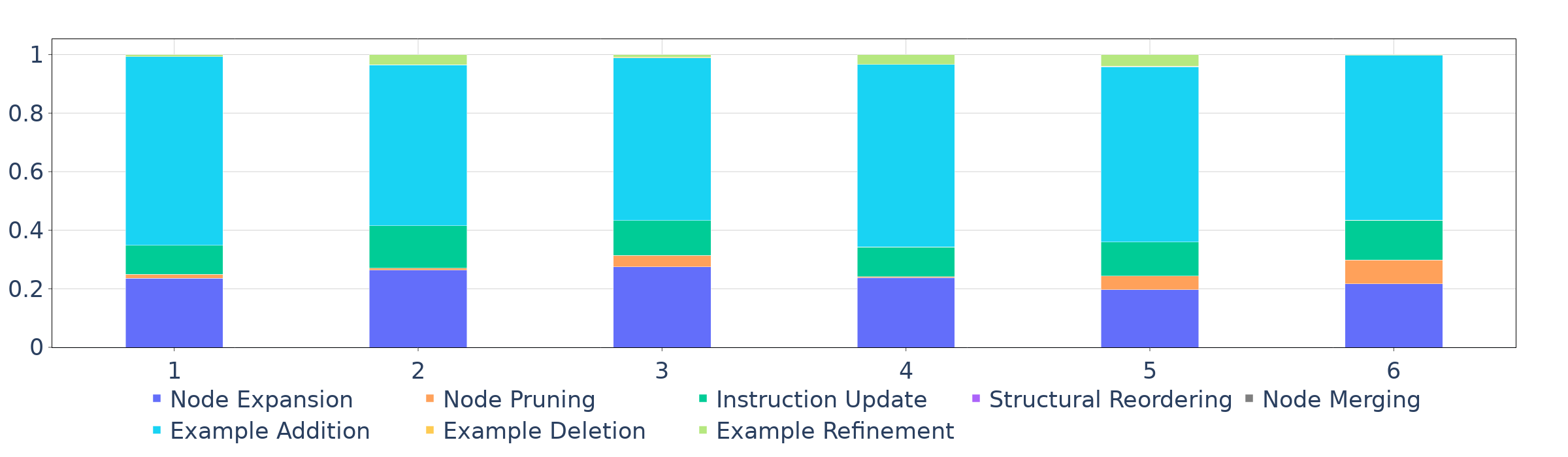}
  \caption{Salient Translation}
\end{subfigure}
\caption{Action distribution over optimization steps in \protegi}
\label{fig:taskwise_protegi_1}
\end{figure*}

\begin{figure*}[!htbp]
\begin{subfigure}{0.99\textwidth}
\centering   \captionsetup{justification=centering} 
  \includegraphics[scale=0.2]{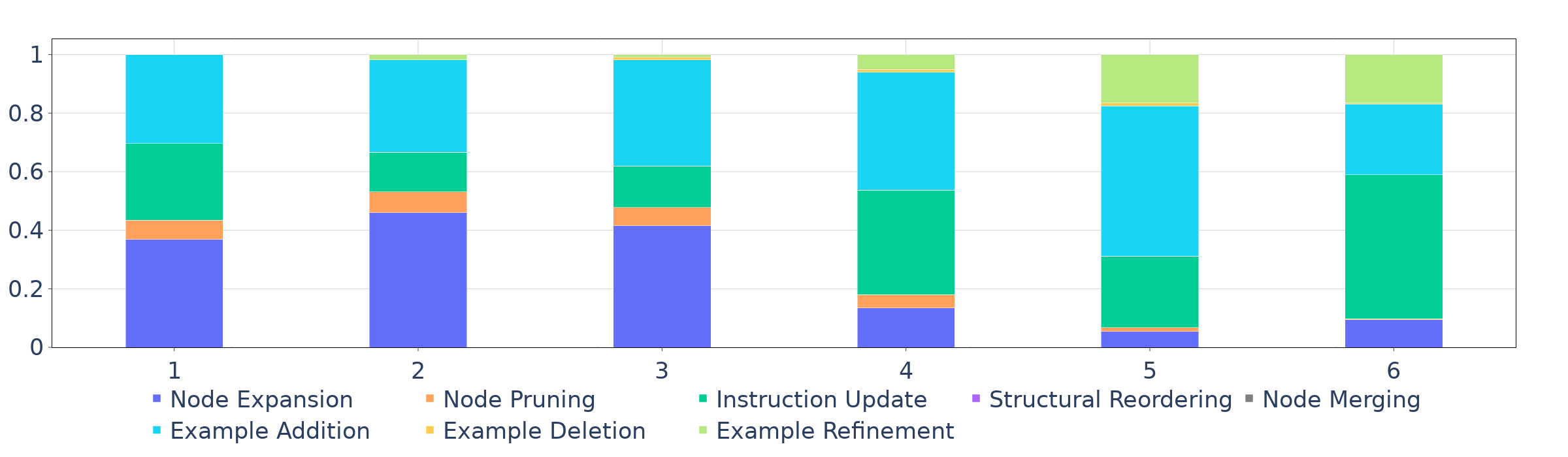}
  \caption{Inappropriate}
\end{subfigure}
\hfill
\begin{subfigure}{0.99\textwidth}
  \centering   \captionsetup{justification=centering} 
  \includegraphics[scale=0.2]{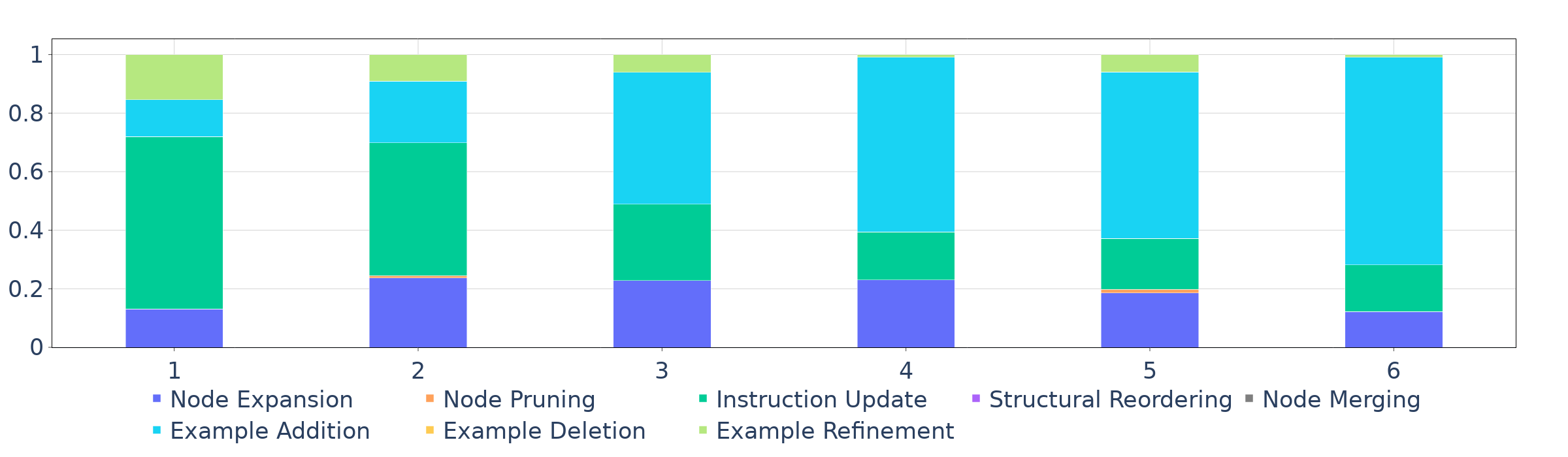}
  \caption{Misinformation}
\end{subfigure}
\hfill
\begin{subfigure}{0.99\textwidth}
  \centering   \captionsetup{justification=centering} 
  \includegraphics[scale=0.2]{images/action_types_across_task_sculpt_Offensive.png}
  \caption{Hate}
\end{subfigure}
\hfill
\begin{subfigure}{0.99\textwidth}
  \centering   \captionsetup{justification=centering} 
  \includegraphics[scale=0.2]{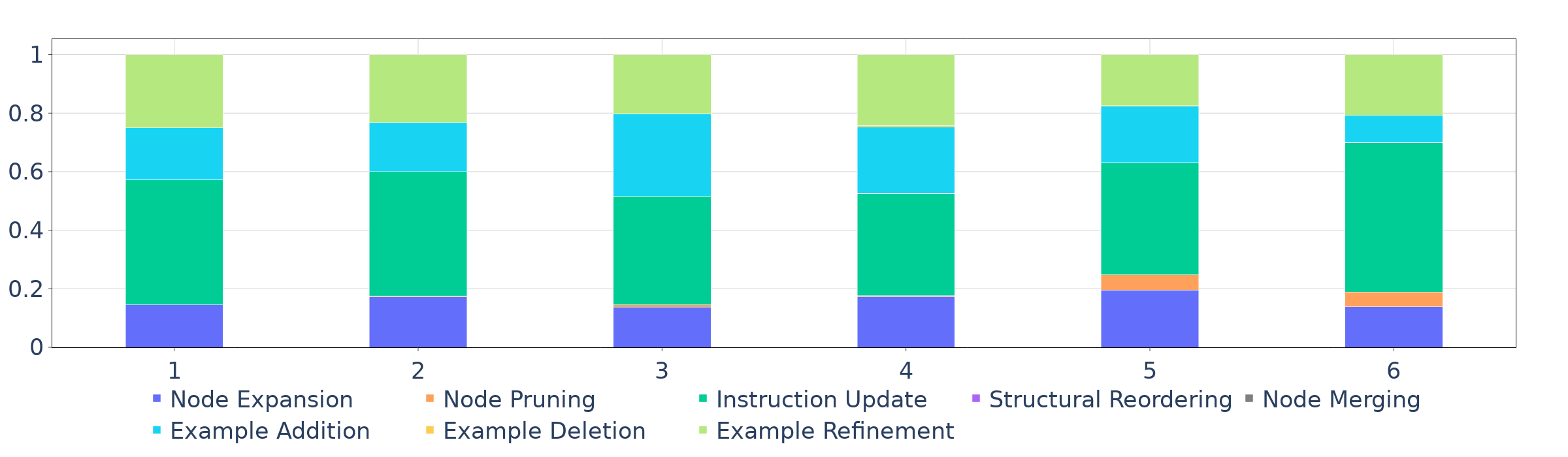}
  \caption{SelfHarm}
\end{subfigure}
\caption{Action distribution over optimization steps in \protegi}
\label{fig:taskwise_protegi_2}
\end{figure*}

Additionally, Figures \ref{fig:taskwise_sculpt_1} and \ref{fig:taskwise_sculpt_2} show the action distribution employed by SCULPT over successive optimization steps, providing insights into how the system refines prompts over time. Similarly, Figures \ref{fig:taskwise_protegi_1} and \ref{fig:taskwise_protegi_2} illustrate the stepwise action distribution for ProTeGi, highlighting differences in how the two methods evolve their optimizations throughout the process.

\begin{table*}[!htbp]
\centering
\begingroup 
\setlength{\parindent}{0pt} 
\renewcommand{\arraystretch}{1.5} 
\begin{tabular}{ll}
\toprule
Prompt Reference           & Preliminary Assessment  \\
\midrule
\begin{minipage}{2.15in}
\vspace{2mm}
\begin{scriptsize}
\begin{spverbatim}
- Task> body",
- Error Identification> body
- Error Identification> 1.> body
- Error Identification> 2.> body
- Error Identification> 3.> body
- Error Identification> 4.> body
- Error Identification> 5.> body
- Error Identification> 6.> body
- Performance Analysis> body
- Additional points> body
- Options> body
- Output format> body
\end{spverbatim}
\end{scriptsize}
\vspace{2mm}
\end{minipage} & 
\begin{minipage}{3.8in}
\vspace{2mm}
\begin{scriptsize}
\begin{spverbatim}
- Add relevant examples to each error class to provide clarity.
- Simplify language where possible to make instructions clearer.
- Combine sections where appropriate to reduce redundancy and improve flow.
\end{spverbatim}
\end{scriptsize}
\vspace{2mm}
\end{minipage}\\ 

\bottomrule
\end{tabular}
\endgroup 
\caption{Example of Preliminary Critic Reflection}
\label{tab:prelim-critic-reponse-ST}
\end{table*}

\begin{table*}[!htbp]
\centering
\begingroup 
\setlength{\parindent}{0pt} 
\begin{tabular}{lll}
\toprule
Action Type             &Section Reference           & Action Description  \\
\midrule
\begin{minipage}{1.4in}
\vspace{2mm}
\begin{scriptsize}
\begin{spverbatim}
Example Update- Addition
\end{spverbatim}
\vspace{2mm}
\end{scriptsize}
\end{minipage} &
\begin{minipage}{1.4in}
\vspace{2mm}
\begin{scriptsize}
\begin{spverbatim}
Error Identification> 1.
\end{spverbatim}
\end{scriptsize}
\end{minipage} & 
\begin{minipage}{3in}
\vspace{2mm}
\begin{scriptsize}
\begin{spverbatim}
Add examples that illustrate 'Named entities' errors, such as changes in names, places, or locations. For instance, 'A person's name changed from John to Jack' or 'A city name changed from New York to Los Angeles'
\end{spverbatim}
\end{scriptsize}
\vspace{2mm}
\end{minipage}\\ 

\hline
\begin{minipage}{1.4in}
\vspace{2mm}
\begin{scriptsize}
\begin{spverbatim}
"Section Rephrase"
\end{spverbatim}
\vspace{2mm}
\end{scriptsize}
\end{minipage} & \begin{minipage}{1.4in}
\vspace{2mm}
\begin{scriptsize}
\begin{spverbatim}
"Task> body"
\end{spverbatim}
\end{scriptsize}
\end{minipage} & 
\begin{minipage}{3in}
\vspace{2mm}
\begin{scriptsize}
\begin{spverbatim}
Your task is to identify the type of translation error in a given source-translation pair. You will be given sentences with specific errors manually introduced. Determine which of the six error classes the translation error belongs to.
\end{spverbatim}
\end{scriptsize}
\vspace{2mm}
\end{minipage}\\ 

\bottomrule
\end{tabular}
\endgroup 
\caption{Actor Responses based on Preliminary Assessment}
\label{tab:prelim-actor-reponse-ST}
\end{table*}

\begin{table*}[!htbp]
\centering
\begingroup 
\setlength{\parindent}{0pt} 
\renewcommand{\arraystretch}{1.5} 
\begin{tabular}{lll}
\toprule
Input Example \& Prediction  & Prompt Reference           &  Error Assessment  \\
\midrule
\begin{minipage}{1.7in}
\vspace{2mm}
\begin{scriptsize}
\begin{spverbatim}
Input: "Source:  Eleonore Lingnau-Kluge war eine deutsche Malerin.
Translation: Eleonore Lingnau-Kluge was a German dancer."
Expected Output: "(F)"
Prediction: "(D)"
\end{spverbatim}
\vspace{2mm}
\end{scriptsize}
\end{minipage} & \begin{minipage}{1.6in}
\vspace{2mm}
\begin{scriptsize}
\begin{spverbatim}
- Error Identification> 1.> body
- Error Identification> 5.> body
- Error Identification> 6.> Examples
\end{spverbatim}
\end{scriptsize}
\end{minipage} & 
\begin{minipage}{2.5in}
\vspace{2mm}
\begin{scriptsize}
\begin{spverbatim}
- Add examples that clearly differentiate factual errors from named entity changes.
- Clarify the definitions of each error class to avoid confusion.
\end{spverbatim}
\end{scriptsize}
\vspace{2mm}
\end{minipage}\\ 

\hline
\begin{minipage}{1.7in}
\vspace{2mm}
\begin{scriptsize}
\begin{spverbatim}
"Input: "Source:  Pedro Morenés y Álvarez de Eulate ist ein spanischer Politiker der Partido Popular.
Translation: Pedro is a Spanish politician of the Popular Party."
Expected Output: "(D)"
Prediction: "(E)"
\end{spverbatim}
\vspace{2mm}
\end{scriptsize}
\end{minipage} & \begin{minipage}{1.6in}
\vspace{2mm}
\begin{scriptsize}
\begin{spverbatim}
- Error Identification> 1.> body
- Error Identification> 6.> Examples
\end{spverbatim}
\end{scriptsize}
\end{minipage} & 
\begin{minipage}{2.5in}
\vspace{2mm}
\begin{scriptsize}
\begin{spverbatim}
- Add examples that highlight named entity changes, especially when names are shortened or altered.",
- Emphasize the importance of preserving named entities in translations.
\end{spverbatim}
\end{scriptsize}
\vspace{2mm}
\end{minipage}\\ 

\bottomrule
\end{tabular}
\endgroup 
\caption{Error Assessment from Critic}
\label{tab:error-critic-reponse-ST}
\end{table*}

\begin{table*}[!htbp]
\centering
\begingroup 
\setlength{\parindent}{0pt} 
\renewcommand{\arraystretch}{1.5} 
\begin{tabular}{lll}
\toprule
Action Type        &Section Reference        & Action Description    \\
\midrule
\begin{minipage}{1.4in}
\vspace{2mm}
\begin{scriptsize}
\begin{spverbatim}
Example Update- Addition
\end{spverbatim}
\vspace{2mm}
\end{scriptsize}
\end{minipage} &
\begin{minipage}{1.4in}
\vspace{2mm}
\begin{scriptsize}
\begin{spverbatim}
Error Identification> 1.> Examples
\end{spverbatim}
\vspace{2mm}
\end{scriptsize}
\end{minipage} &
\begin{minipage}{3in}
\vspace{2mm}
\begin{scriptsize}
\begin{spverbatim}
Add examples that highlight named entity changes, especially when names are shortened or altered. For instance: 'A politician's name changed from Pedro Morenés y Álvarez de Eulate to Pedro' and 'An actor's name changed from Martin Stephen McCann to McCann'.
\end{spverbatim}
\end{scriptsize}
\vspace{2mm}
\end{minipage}\\ 

\hline
\begin{minipage}{1.4in}
\vspace{2mm}
\begin{scriptsize}
\begin{spverbatim}
Section Rephrase
\end{spverbatim}
\vspace{2mm}
\end{scriptsize}
\end{minipage} & 
\begin{minipage}{1.4in}
\vspace{2mm}
\begin{scriptsize}
\begin{spverbatim}
Error Identification> 1.> body
\end{spverbatim}
\end{scriptsize}
\vspace{2mm}
\end{minipage}& 
\begin{minipage}{3in}
\vspace{2mm}
\begin{scriptsize}
\begin{spverbatim}
Named entities: Look for changes in names, places, locations, etc. Ensure that names are preserved accurately, even when shortened or altered.
\end{spverbatim}
\end{scriptsize}
\vspace{2mm}
\end{minipage}\\ 

\bottomrule
\end{tabular}
\endgroup 
\caption{Actor Response based on Error Assessment}
\label{tab:error-assess-actor-reponse-ST}
\end{table*}

\begin{table*}[!htbp]
\centering
\begingroup 
\setlength{\parindent}{0pt} 
\renewcommand{\arraystretch}{1.5} 
\begin{tabular}{|l|}
\hline
\begin{minipage}{6.1in}
\vspace{2mm}
\begin{scriptsize}
\begin{lstlisting}[style=mycustomstyle]
# Task
(*@\rephrasedPrompt{Your task is to identify the type of translation error in a given source-translation pair. You will be given sentences with}@*) 
(*@\rephrasedPrompt{specific errors manually introduced. Determine which of the six error classes the translation error belongs to.}@*)

# Error Identification
Analyze the source-translation pair and identify the error based on the following classes:
* Named entities: Look for changes in names, places, locations, etc.
(*@\examplePrompt{Examples: {A company name changed from Apple to Microsoft}, {A country name changed from France to Germany}, {A person's name changed from John to Jack}, {A city name changed from New York to Los Angeles}}@*)
* Numerical values: Check for alterations in numbers, dates, or units.
(*@\examplePrompt{Examples: {A date changed from 2021 to 2020}, {A price changed from \$50 to \$55}, {A time changed from 3 PM to 4 PM}, {A measurement unit changed from meters to feet}, {A quantity changed from 100 to 150}}@*)
* Modifiers or adjectives: Identify changes in descriptors pertaining to a noun.
(*@\examplePrompt{Examples: {The adjective changed from big to small.}, {The descriptor changed from red to blue.}, {The modifier changed from happy to sad.}, {The descriptor changed from old to new.}, {The adjective changed from tall to short.}}@*)
* Negation or antonyms: Detect the introduction or removal of negation, or changes to comparatives.
(*@\examplePrompt{Examples: {The comparative changed from 'less important' to 'more important'.}, {The negation changed from 'is not' to 'is'.}, {The phrase changed from 'He is not happy' to 'He is happy'.}, {The comparative changed from 'better' to 'worse'.}, {The sentence changed from 'She never goes to the gym' to 'She always goes to the gym'.}}@*)
* Facts: Spot trivial factual errors not covered by the above classes.
(*@\examplePrompt{Examples: {The fact changed from The capital of France is Paris to The capital of France is Berlin}, {The fact changed from Humans have 206 bones in their body to Humans have 210 bones in their body}, {The fact changed from The Great Wall of China is visible from space to The Great Wall of China is not visible from space}}@*)
* Dropped content: Notice if a significant clause is missing from the translation.
Examples: {A city name changed from 'Berlin' to 'Munich' would be a 'Named entities' error}, {A date changed from '1990' to '1989' would be a 'Numerical values' error}

# Performance Analysis
(*@\changeprompt{Understand that language models perform differently across error classes:}@*)
* Models like XLM-Roberta may struggle with named entities, dropped content, and modifiers/adjectives.
* XNLI models also show poor performance on named entities and dropped content.

# Additional points
(*@\changeprompt{Keep in mind the following points while identifying errors:}@*)
* Ensure minimal impact on translation fluency while identifying errors.
* Focus on salient source information to detect errors effectively.
* Remember that each translation contains only one of the six error classes.

# Options
(A) Modifiers or Adjectives 
(B) Numerical Values 
(C) Negation or Antonyms 
(D) Named Entities 
(E) Dropped Content 
(F) Facts

# Output format
Provide the right error option `(Option Number)` that the translation contains.
\end{lstlisting}
\end{scriptsize}
\vspace{2mm}
\end{minipage}\\ 

\hline
\end{tabular}
\endgroup 
\caption{Updated Prompt after Preliminary Assessment. Instruction update are highlighted in \rephrasedPrompt{blue}, Example Addition are marked with \examplePrompt{mahogany} color, and Node Expansion are marked in \changeprompt{green}.}
\label{tab:prompt-after-prelim}
\end{table*}

\begin{table*}[!htbp]
\centering
\begingroup 
\setlength{\parindent}{0pt} 
\renewcommand{\arraystretch}{1.5} 
\begin{tabular}{|l|}
\hline
\begin{minipage}{6.1in}
\vspace{2mm}
\begin{scriptsize}
\begin{lstlisting}[style=mycustomstyle]
# Task
(*@\rephrasedPrompt{Your task is to identify the type of translation error in a given source-translation pair. You will be given sentences with}@*) 
(*@\rephrasedPrompt{specific errors manually introduced. Determine which of the six error classes the translation error belongs to.}@*)

# Error Identification
Analyze the source-translation pair and identify the error based on the following classes:
* Named entities: Look for changes in names, places, locations, etc. Ensure that names are preserved accurately, even when shortened or altered.
(*@\examplePrompt{Examples: {A politician's name changed from Pedro Morenés y Álvarez de Eulate to Pedro}, {A company name changed from Apple to Microsoft}, {An actor's name changed from Martin Stephen McCann to McCann}, {A country name changed from France to Germany}, {A person's name changed from John to Jack}, {A city name changed from New York to Los Angeles}}@*)
* Numerical values: Check for alterations in numbers, dates, or units. Ensure that numerical values are preserved accurately.
(*@\examplePrompt{Examples: {A date changed from 2021 to 2020}, {A price changed from \$50 to \$55}, {A time changed from 3 PM to 4 PM}, {The population number changed from 5491 to 5000}, {A measurement unit changed from meters to feet}, {A quantity changed from 100 to 150}}@*)
* Modifiers or adjectives: Identify changes in descriptors pertaining to a noun. Ensure that descriptive terms are preserved accurately.
(*@\examplePrompt{Examples: {The adjective changed from big to small.}, {The descriptor changed from red to blue.}, {The modifier changed from happy to sad.}, {The description changed from small town to city.}, {The descriptor changed from old to new.}, {The nationality changed from German to French.}, {The adjective changed from tall to short.}}@*)
(*@\rephrasedPrompt{* Negation or antonyms: Detect the introduction or removal of negation, or changes to comparatives. Ensure that negation and antonyms are preserved accurately.}@*)
(*@\examplePrompt{Examples: {The comparative changed from 'less important' to 'more important'.}, {The negation changed from 'is not' to 'is'.}, {The description changed from rural to urban.}, {The phrase changed from 'He is not happy' to 'He is happy'.}, {The comparative changed from 'better' to 'worse'.}, {The sentence changed from 'She never goes to the gym' to 'She always goes to the gym'.}}@*)
* Facts: Spot trivial factual errors not covered by the above classes. Ensure that factual information remains accurate and unchanged.
(*@\examplePrompt{Examples: {The fact changed from The Pacific Ocean is the largest ocean to The Atlantic Ocean is the largest ocean}, {The fact changed from The Earth orbits the Sun to The Sun orbits the Earth}, {The fact changed from The Great Wall of China is visible from space to The Great Wall of China is not visible from space}, {The fact changed from The capital of France is Paris to The capital of France is Berlin}, {The fact changed from Humans have 206 bones in their body to Humans have 210 bones in their body}}@*)
* Dropped content: Notice if a significant clause is missing from the translation.
(*@\examplePrompt{Examples: {A city name changed from 'Berlin' to 'Munich' would be a 'Named entities' error}, {A date changed from '1990' to '1989' would be a 'Numerical values' error}}@*)

# Performance Analysis
(*@\changeprompt{Understand that language models perform differently across error classes:}@*)
* Models like XLM-Roberta may struggle with named entities, dropped content, and modifiers/adjectives.
* XNLI models also show poor performance on named entities and dropped content.

# Additional points
(*@\changeprompt{Keep in mind the following points while identifying errors:}@*)
* Ensure minimal impact on translation fluency while identifying errors.
* Focus on salient source information to detect errors effectively.
* Remember that each translation contains only one of the six error classes.

# Options
(A) Modifiers or Adjectives
(B) Numerical Values
(C) Negation or Antonyms
(D) Named Entities
(E) Dropped Content
(F) Fact

# Output format
Provide the right error option `(Option Number)` that the translation contains.
\end{lstlisting}
\end{scriptsize}
\vspace{2mm}
\end{minipage}\\ 

\hline
\end{tabular}
\endgroup 
\caption{Updated Prompt after  Error Assessment. Instruction update are highlighted in \rephrasedPrompt{blue}, Example Addition are marked with \examplePrompt{mahogany} color, and Node Expansion are marked in \changeprompt{green}.}
\label{tab:prompt-after-error-assess}
\end{table*}

\begin{table*}[!htbp]
\centering
\begingroup 
\setlength{\parindent}{0pt} 
\renewcommand{\arraystretch}{1.5} 
\begin{tabular}{|l|}
\hline
\begin{minipage}{6.1in}
\vspace{2mm}
\begin{scriptsize}
\begin{lstlisting}[style=mycustomstyle]
# Task
(*@\rephrasedPrompt{Your task is to identify the type of translation error in a given source-translation pair. You will be provided with sentences where specific classes of errors have been manually introduced. Determine which of the six error classes the translation error belongs to: Named entities, Numerical values, Modifiers or adjectives, Negation or antonyms, Facts, and Dropped content.}@*)
(*@\examplePrompt{Examples: {The name 'John' was changed to 'James' in the translation, which is a 'Named entities' error.}, {The word 'happy' was translated as 'sad', which is a 'Negation or antonyms' error.}, {The number '50' was translated as '15', which is a 'Numerical values' error.}}@*)

# Error Identification
Analyze the provided source-translation pair and identify the error based on the following classes:
(*@\rephrasedPrompt{* Named entities: Look for changes in names, places, locations, scientific names, classifications, etc. This includes any change to a name, including shortening, omission, or alteration of specific locations. Pay attention to changes in classifications that might alter the meaning or context of the sentence.}@*)
(*@\examplePrompt{Examples: {The name 'New York' was changed to 'NY', indicating a 'Named entities' error.}, {The phrase 'United States' was modified to 'USA', indicating a 'Named entities' error.}, {The term 'California' was altered to 'CA', indicating a 'Named entities' error.}, {The name 'Boyd Kevin Rutherford' was reduced to 'Boyd' in the translation, indicating a 'Named entities' error.}, {The term 'Rabenvogel' was incorrectly translated as 'Columbine family', changing the classification.}}@*)
(*@\rephrasedPrompt{* Numerical values: Check for alterations in numbers, dates, or units, and ensure that no numerical information is omitted. This includes any change, omission, or alteration of numerical data. Pay attention to omissions that might alter the meaning or context of the sentence.}@*)
(*@\examplePrompt{Examples: {The date '2021' was omitted.}, {The number '100' was changed to 'one hundred'.}, {The unit 'kg' was altered to 'kilogram'.}, {The dates were omitted, losing important context.}, {The population '5491' was omitted, which is a numerical value.}}@*)
(*@\rephrasedPrompt{* Modifiers or adjectives: Identify changes in descriptors pertaining to a noun that are not necessarily antonyms. This includes changes in descriptors such as nationality, type, usage, or any other descriptive attribute. Pay attention to changes that might alter the meaning or context of the sentence.}@*)
(*@\examplePrompt{Examples: {The adjective 'quick' was changed to 'speedy' in the report.}, {The term 'Rosenmontagszug' was translated as 'Rose Procession', changing the descriptor.}, {The phrase 'modern' was altered to 'contemporary' in the article.}, {The adjective 'happy' was changed to 'joyful' in the sentence.}}@*)
(*@\rephrasedPrompt{* Negation or antonyms: Detect the introduction or removal of negation, or changes to comparatives. This includes any change that introduces or removes a negative meaning or alters the comparative degree of an adjective or adverb. Pay attention to antonyms that might alter the meaning or context of the sentence.}@*)
(*@\examplePrompt{Examples: {Changing 'more important' to 'less important' is a comparative change.}, {Changing 'Obere' to 'Lower' is an antonym.}, {Changing 'living' to 'extinct' is an antonym.}, {Changing 'He is not interested' to 'He is interested' would be a 'Negation or antonyms' error.}, {Changing 'better' to 'worse' is a comparative change.}}@*)
(*@\rephrasedPrompt{* Facts: Spot trivial factual errors not covered by the above classes. This includes changes to factual information such as professions. Pay attention to errors that might alter the factual accuracy of the sentence.}@*)
(*@\examplePrompt{Examples: {Asserting that 'Neil Armstrong was the first person to climb Mount Everest' instead of 'Neil Armstrong was the first person to walk on the moon' is a 'Facts' error.}, {Saying 'The Great Wall of China is located in India' instead of 'The Great Wall of China is located in China' is a 'Facts' error.}, {Stating that 'The capital of France is Berlin' instead of 'The capital of France is Paris' is a 'Facts' error.}, {Claiming that 'Albert Einstein was a famous painter' instead of 'Albert Einstein was a famous physicist' is a 'Facts' error.}, {Stating that 'The Amazon River is the longest river in the world' instead of 'The Nile River is the longest river in the world' is a 'Facts' error.}}@*)
(*@\rephrasedPrompt{* Dropped content: Identify if a significant clause or important information is missing from the translation. Pay attention to omissions that might alter the meaning or context of the sentence.}@*)
(*@\examplePrompt{Examples: {The clause 'which is located in the heart of the city' is omitted, losing important location context.}, {The phrase 'including taxes' is omitted, which is crucial for understanding the total cost.}, {The information 'who is a renowned scientist' is missing, which provides important context about the individual.}}@*)

(*@\st{\# Performance Analysis}@*)
(*@\st{Understand that existing language models have varying performance across different error classes:}@*)
(*@\st{* Models like XLM-Roberta may struggle with named entities, dropped content, and modifiers/adjectives.}@*)
(*@\st{* XNLI models also show poor performance on named entities and dropped content.}@*)

(*@\st{\# Additional points}@*)
(*@\st{* Ensure minimal impact on translation fluency while identifying errors.}@*)
(*@\st{* Focus on salient source information to detect errors effectively.}@*)
(*@\st{* Remember that each translation contains only one of the six error classes.}@*)

# Options
(A) Modifiers or Adjectives 
(B) Numerical Values 
(C) Negation or Antonyms 
(D) Named Entities 
(E) Dropped Content 
(F) Facts

(*@\changeprompt{\#\# Options Explanation}@*)
(*@\changeprompt{Explanation of each option: }@*)
(*@\changeprompt{(A) Modifiers or Adjectives: Changes in descriptors pertaining to a noun. }@*)
(*@\changeprompt{(B) Numerical Values: Alterations in numbers, dates, or units.}@*) 
(*@\changeprompt{(C) Negation or Antonyms: Introduction or removal of negation, or changes to comparatives. }@*)
(*@\changeprompt{(D) Named Entities: Changes in names, places, locations, etc. }@*)
(*@\changeprompt{(E) Dropped Content: Missing significant clauses from the translation. }@*)
(*@\changeprompt{(F) Facts: Trivial factual errors not covered by the above classes.}@*)

# Output format
(*@\rephrasedPrompt{Provide the correct error option (A-F) that the translation contains.}@*)
\end{lstlisting}
\end{scriptsize}
\vspace{2mm}
\end{minipage}\\ 

\hline
\end{tabular}
\endgroup 
\caption{Updated Prompt at the end of Optimization process. Instruction update are highlighted in \rephrasedPrompt{blue}, Example Addition are marked with \examplePrompt{mahogany} color, \st{strike through} indicates Node Pruning, and Node Expansion are marked in \changeprompt{green}.}

\label{tab:prompt-after-process-completion}
\end{table*}

\clearpage

\section{APE Template}
\label{sec:ape}
\noindent\textbf{Forward Generation}
\begin{scriptsize}
\begin{spverbatim}
I gave a friend an instruction and {NumExamples} inputs. The friend read the instruction and wrote an output for every one of the inputs. Given, the input-output pairs, generate an instruction which is the output. Generate the output between the <INSTRUCT> and <ENDINSTRUCT> Tags.
\end{spverbatim}
\end{scriptsize}

\vspace{1em}
\noindent\textbf{Reverse Generation}
\begin{scriptsize}
\begin{spverbatim}
I instructed my friend to <INSERT>. The friend read the instruction and wrote an output for every one of the inputs. Given the input and output pairs, complete the <INSERT> instruction. Generate the output between the <INSTRUCT> and <ENDINSTRUCT> Tags.
\end{spverbatim}
\end{scriptsize}

\section{LAPE Template}
\label{sec:long_ape}
\noindent\textbf{Forward Generation}
\vspace{1em}
\begin{scriptsize}
\begin{spverbatim}
I gave a friend a detailed instruction in markdown format and {NumExamples} inputs. The instructions has the following markdown structure with proper white spaces-

```
# <Heading 1>
<body>

## <Heading 1.1>
<body>
Examples: {example 1}, {example 2}

* <bullet point 1>
* <bullet point 2>
Examples: {example 1}, {example 2}, {example 3}
* <bullet point 3>

...

# <Heading 2>
* <bullet point 1>
* <bullet point 2>
...
```

The friend read the instruction and wrote an output for every one of the inputs. The instruction had several sections, each describing what output to generate for a given input. Each section also has examples to assist my friend. 

Given the input-output pairs, generate an instruction which is the output. For each section, you can either use the same input-output pairs to write relevant examples, or you can use your best knowledge to create examples according to the observed input-output pairs. Do not use the input-output pairs directly as provided, you have to maintain the structure of the instruction intact with added examples by keeping each input text in its own in curly brackets and each curly bracketed example separated by comma, and ignore the output if the section describes labelling condition for given output label. Ensure that proper line separation is maintained for readability. Do not reproduce the tags like <body>, <bullet point 1> etc, those represent placeholder for relevant content in the instruction. Generate the output between the <INSTRUCT> and <ENDINSTRUCT> Tags.
\end{spverbatim}
\end{scriptsize}

\vspace{1em}
\noindent\textbf{Reverse Generation}
\vspace{1em}

\begin{scriptsize}
\begin{spverbatim}
I instructed my friend to <INSERT>. The instructions looked something like this-

```
# <Heading 1>
<body>

## <Heading 1.1>
<body>
Examples: {example 1}, {example 2}

* <bullet point 1>
* <bullet point 2>
Examples: {example 1}, {example 2}, {example 3}
* <bullet point 3>

...

# <Heading 2>
* <bullet point 1>
* <bullet point 2>
...
```
The friend read the instruction and wrote an output for every one of the inputs. Given the input and output pairs, complete the <INSERT> instruction.For each section, you can either use the same input-output pairs to write relevant examples, or you can use your best knowledge to create examples according to the observed input-output pairs. Do not use the input-output pairs directly as provided, you have to maintain the structure of the instruction intact with added examples by keeping each input text in its own in curly brackets and each curly bracketed example separated by comma, and ignore the output if the section describes labelling condition for given output label. Ensure that proper line separation is maintained for readability. Do not reproduce the tags like <body>, <bullet point 1> etc, those represent placeholder for relevant content in the instruction. Generate the output between the <INSTRUCT> and <ENDINSTRUCT> Tags.
\end{spverbatim}
\end{scriptsize}
\vspace{1em}

\section{SCULPT Prompt Templates for Critic and Actor}
\label{sec:sculpt_prompts}
In this section, we present the prompt templates that were utilized for generating the Critic and Actor responses using a large language model (LLM). These templates serve as the foundation for eliciting structured feedback from the Critic and actionable suggestions from the Actor during the iterative prompt optimization process in SCULPT.

\vspace{1em}

\subsection{Critic Template for Preliminary Assessment}
\vspace{1em}

\begin{scriptsize}
\begin{spverbatim}
## Step-by-Step Instructions:

1. **Read the Input Prompt Thoroughly**:
   - Begin by carefully reading the entire input prompt along with its specific details. Make sure to understand the task at hand, including any requirements or constraints provided.

2. **General Feedback**:
    Provide comprehensive feedback on the input prompt to enhance its effectiveness in each of the following areas:
    * Contextual Errors: Identify specific inaccuracies or mistakes that may lead to misunderstandings.
    * Incorrect or Irrelevant Examples: Highlight any incorrect or misplaced examples within the prompt. Note that no section should contain more than 5-6 examples.
    * Gaps in Information: Point out any missing details or context that could clarify the task for the user, ensuring they have all necessary information.
    * Potential Improvements: Suggest ways to improve the prompt for better clarity and impact. This could include simplifying language, adding relevant examples, or outlining a clear sequence of steps. Ensure the prompt is efficient, concise, and free from redundant information.
    * Grammar and Syntax: Note any spelling or grammatical errors that could cause confusion, as well as poorly constructed sentences that may obscure the intended meaning.
    * Prompt Length: The prompt should be concise. Provide feedback around optimizing its length while maintaining clarity.
    * Other Issues: Identify any other areas where the prompt could be improved.

## Input Format:

**Example Prompt**

```json
{"<Heading 1>":{"body": "<body>","<Heading 1.1>":{"body": "<body>",...},"<Heading 1.2>":{"body": "<body>","Examples":["<example 1>",....],"<Heading 1.2.1>":{"body": "<body>","1.": {"body": "<instruction>","Examples":["<example 1>","<example 2>",.....]},"2.":...},"<Heading 1.2.2>":{"body": "<body>"}...}...},"<Heading 2>":{"body": "<body>"}...}
```

## Output Format:

```json
{"prompt_feedback": [{"prompt_examination":"<prompt_examination>", "improvement_suggestion": ["<improvement_suggestion>", ...]}, ...], "prompt_references": ["<prompt_reference>", ...]}
```
\end{spverbatim}
\end{scriptsize}
\vspace{1em}

\subsection{Critic Template for Error Assessment}
\vspace{1em}

\begin{scriptsize}
\begin{spverbatim}
# Task
Evaluate the performance of the input prompt and provide explanations, identify the parts of the prompt used for predictions, and offer feedback for improvement.

## Step-by-Step Instructions:

1. **Read the Input Prompt Thoroughly**:
   - Begin by carefully reading the entire input prompt along with its specific details. Make sure to understand the task at hand, including any requirements or constraints provided.

2. **Batch Feedback**:
   Based on `Batch Evaluations` where the model has generated wrong predictions, provide feedback to improve the performance of the prompt by following these steps:
    * Understanding the Context (prediction_explanation):
       - Start by clearly stating the input, expected output (ground_truth), and model’s actual prediction. Example format: `Input: '<input text>'`, `Expected Output: '<ground_truth>'`, `Prediction: '<prediction>'`.
       - Analyze why the model generated this prediction by identifying specific words, phrases, or contextual cues from the input.
       - Highlight the sections of the input or prompt that likely influenced the prediction using `prompt_references`.
               
    * Analysis and Feedback (prompt_feedback):
      The feedback should include the details about each of these steps:
      - `prediction_analysis`: Always include a clear analysis comparing the model's prediction with the expected output. Mention what the correct label should have been and highlight any discrepancies.
      - `prompt_examination`: Always analyze the prompt step-by-step, identifying specific sections that may have caused the error (e.g., unclear instructions, ambiguous wording). Explain how these issues led to the incorrect label.
      - `improvement_suggestions`: Provide multi-step feedback outlining all possible actions to address identified issues and explain how these changes will result in the correct label. Possible actions can include:
         - Rephrasing unclear instructions.
         - Removing redundancy.
         - Adding clarity or details.
         - Revising tone or structure for better flow.
         - Modifying examples: Remove bad examples, add or refine better examples, ensuring no section exceeds 5-6 examples.
         - 
    * Prompt references (`prompt_references`):
      Include references to the specific parts of the prompt that may have contributed to errors.
      
## Input Format:

**Example Prompt**

```json
{"<Heading 1>":{"body": "<body>","<Heading 1.1>":{"body": "<body>",...},"<Heading 1.2>":{"body": "<body>","Examples":["<example 1>",....],"<Heading 1.2.1>":{"body": "<body>","1.": {"body": "<instruction>","Examples":["<example 1>","<example 2>",.....]},"2.":...},"<Heading 1.2.2>":{"body": "<body>"}...}...},"<Heading 2>":{"body": "<body>"}...}
```

**Example Batch Evaluations**

```json
{"prompt": "The current prompt being used.","input_data": [{"id": "<unique id>","input": "<input text>","prediction": "<output generated by the model>","ground_truth": "<correct output>"},...]}
```

## Output Format:

```json
[{"id": "<unique id>","prediction_explanation": "<explanation for prediction>","prompt_feedback": {"prediction_analysis": "<prediction_analysis>", "prompt_examination": "<prompt_examination>", "improvement_suggestions": ["<improvement_suggestions>", ...]},"prompt_references": ["Heading 1> Heading 1.2> Heading 1.2.1> body","Heading 1> Heading 1.2> Heading 1.2.1> 2.> body","Heading 1> Heading 1.2> body","Heading 2> body>"]},...]
```
\end{spverbatim}
\end{scriptsize}
\vspace{1em}

\subsection{Critic Template for Error Assessment using Similarity-driven Aggregation}
\vspace{1em}

\begin{scriptsize}
\begin{spverbatim}
# Task Overview  
Evaluate a set of prompts, predictions, and ground truths. Provide detailed feedback on each case and group related feedback into clusters based on common patterns or prompt references.

## Step-by-Step Instructions:

1. **Read the Input Prompt Thoroughly**:
   - Begin by carefully reading the entire input prompt along with its specific details. Make sure to understand the task at hand, including any requirements or constraints provided.

2. **Batch Feedback**:
   Based on `Batch Evaluations` where the model has generated wrong predictions, provide feedback to improve the performance of the prompt by following these steps:
    * Understanding the Context (prediction_explanation):
       - Start by clearly stating the input, expected output (ground_truth), and model’s actual prediction. Example format: `Input: '<input text>'`, `Expected Output: '<ground_truth>'`, `Prediction: '<prediction>'`.
       - Analyze why the model generated this prediction by identifying specific words, phrases, or contextual cues from the input.
       - Highlight the sections of the input or prompt that likely influenced the prediction using `prompt_references`.
    
    * Analysis and Feedback (`prompt_feedback`):
      The feedback should include the details about each of these steps:
      - `prediction_analysis`: Always include a clear analysis comparing the model's prediction with the expected output. Mention what the correct label should have been and highlight any discrepancies.
      - `prompt_examination`: Always analyze the prompt step-by-step, identifying specific sections that may have caused the error (e.g., unclear instructions, ambiguous wording). Explain how these issues led to the incorrect label.
      - `improvement_suggestions`: Provide multi-step feedback outlining all possible actions to address identified issues and explain how these changes will result in the correct label. Possible actions can include:
         - Rephrasing unclear instructions.
         - Removing redundancy.
         - Adding clarity or details.
         - Revising tone or structure for better flow.
         - Modifying examples: Remove bad examples, add or refine better examples, ensuring no section exceeds 5-6 examples.
    
    * Prompt references (`prompt_references`):
      Include references to the specific parts of the prompt that may have contributed to errors.
    
    * Cluster Feedback:
       - Group related feedback into {number_of_clusters} clusters based on patterns such as:
         - Shared sections of the prompt that influenced the predictions.
         - Expected output `ground_truth`.
         - Similar types of input data or prediction behavior.
       - Each cluster should include:
         - A list of explanations for the inputs in the cluster.
         - A specific list of feedback relevant to the cluster.
         - Clear `prompt_references` pointing to sections that could be revised or improved.

## Input Format:

**Example Prompt**

```json
{"<Heading 1>":{"body": "<body>","<Heading 1.1>":{"body": "<body>",...},"<Heading 1.2>":{"body": "<body>","Examples":["<example 1>",....],"<Heading 1.2.1>":{"body": "<body>","1.": {"body": "<instruction>","Examples":["<example 1>","<example 2>",.....]},"2.":...},"<Heading 1.2.2>":{"body": "<body>"}...}...},"<Heading 2>":{"body": "<body>"}...}
```

**Example Batch Evaluations**

```json
{"prompt": "The current prompt being used.","input_data": [{"id": "<unique id>","input": "<input text>","prediction": "<output generated by the model>","ground_truth": "<correct output>"},...]}
```

## Output Format:
The output must consists of a list of maximum {number_of_clusters} clusters, each identified by a unique `id`.

```json
[{"id": "cluster_1","prediction_explanation": ["<detailed explanation for example 1 in cluster 1>",...], "prompt_feedback": {"prediction_analysis": "<prediction_analysis>", "prompt_examination": "<prompt_examination>", "improvement_suggestions": ["<improvement_suggestions for cluster 1>", ...]}, "prompt_references": ["Heading 1> Heading 1.2> Heading 1.2.1> body","Heading 1> Heading 1.2> Heading 1.2.1> 2.> body","Heading 1> Heading 1.2> body","Heading 2> body"]},...]
```
\end{spverbatim}
\end{scriptsize}
\vspace{1em}

\subsection{Actor Module Template}
\vspace{1em}

\begin{scriptsize}
\begin{spverbatim}
# Task
Use the provided critic feedback to enhance the effectiveness of a prompt. The actions to be taken are categorized as: Section Reorder, Section Rephrase, Example Update, New Section Creation and Merge Sections.

# Example Prompt Structure  
```json  
{"<Heading 1>":{"body": "<body>","<Heading 1.1>":{"body": "<body>",...},"<Heading 1.2>":{"body": "<body>","Examples":["<example 1>",....],"<Heading 1.2.1>":{"body": "<body>","1.": {"body": "<instruction>","Examples":["<example 1>","<example 2>",.....]},"2.":...},"<Heading 1.2.2>":{"body": "<body>"}...}...},"<Heading 2>":{"body": "<body>"}...}
```

## Step-by-Step Instructions for Enhancing a Prompt  

1. **Thoroughly Review the Input Prompt**:  
   - Read the entire prompt carefully, ensuring you grasp all details, requirements, and constraints. Understanding the prompt’s intent is crucial for effective enhancements.

2. **Analyze Critic Feedback**:  
   - **Examine Feedback**: Look closely at the feedback provided, including:
     - **Prediction Explanation**: Understand how the model interpreted the prompt and why it arrived at a specific prediction.
     - **Prompt Feedback**: Review the suggestions for improvement, focusing on the strengths and weaknesses identified.
   - **Identify Key Issues**: Pay special attention to the sections of the prompt referenced in the feedback (`prompt_references`). Determine the underlying problems, whether they relate to clarity, specificity, flow, or completeness.

3. **Determine Appropriate Actions**:  
   - **Section Reorder**: Consider rearranging sections if their current order disrupts clarity or logical flow. Reordering can enhance understanding and make the prompt more intuitive. **Note**: Just the `body` or `Examples` cannot be reordered. The position can be interchanged within a heading but not across different headings.
   - **Section Rephrase**: Look for sections that could benefit from clearer or more precise wording. Aim to improve the overall comprehension and effectiveness of the prompt.
   - **Example Update**: Assess the examples provided. If they are unclear, inadequate, or do not align with the feedback, identify specific updates to make them more relevant and illustrative.
     - Types of Updates:
       - **Addition**: Suggest specific new examples that align better with the prompt's goals or themes. Clearly describe what the new examples should illustrate. **Note**: Ensure that any section does not contain more than 5-6 examples.
       - **Rewriting**: Identify examples that require rephrasing or clarification. Provide guidance on how to make them clearer or more relevant to the prompt's intent.
       - **Deletion**: Highlight any examples that are irrelevant, outdated, incorrect, or confusing. Explain why they should be removed to enhance the clarity of the prompt.
   - **Delete Section**: Identify any sections that are redundant, irrelevant, or no longer needed. Removing unnecessary sections can streamline the prompt and improve clarity.
   - **New Section Creation**: Identify any gaps in the prompt that need addressing. Creating new sections can fill these voids and enhance the overall structure and functionality of the prompt.
   - **Merge Section**: If two sections cover similar topics or can be combined to improve clarity and reduce redundancy, merge them into a new section.

4. **Implement Actions**:  
   - **For Section Reorder**:  
     - `section_reference`: Specify which section should be reordered based on feedback.  
     - `new_position`: Indicate where this section should be moved to improve flow.  
     - `action_explanation`: Explain how this reordering addresses the feedback and enhances prompt clarity.  
   
   - **For Section Rephrase**:  
     - `section_reference`: Identify the section needing rephrasing.  
     - `updated_section`: Provide the revised wording for that section.  
       - `key`: The updated title or heading.  
       - `value`: The rephrased content.  
     - `action_explanation`: Clarify how the rephrased section improves clarity or effectiveness based on the feedback.  
   
   - **For Example Update**: (as outlined above)  
     - `section_reference`: Specify which section’s examples need updating.  
     - `update_type`: Include details on adding, revising, or removing examples.  
     - `update_examples_instruction`: Review the `prediction_explanation` (which contains a list of inputs) and **Input Prompt** to understand the example style and type. Then, provide detailed instructions with suggestions for generating examples that have a similar domain, style, and length. **Reminder**: No section should have more than 5-6 examples.
     - `action_explanation`: Justify the updates based on feedback.  
    
   - **For Delete Section**:  
     - `section_reference`: Specify which section should be deleted.  
     - `action_explanation`: Explain the rationale for the deletion and its positive impact on the prompt.  
   
   - **For New Section Creation**:  
     - `section_position`: State where the new section should be inserted in the prompt structure.  
     - `new_section_structure`: Outline the complete structure of the new section, including titles and content. **Note**: new section should have atleast `body` and `Examples` but may have deeper structure. 
     - `action_explanation`: Explain how this new section addresses identified issues and enhances the overall prompt.
    
   - **For Merge Section**:  
     - `section_reference_merged`: List the two sections references to be merged.
     - `section_position`: State where the merged section should be inserted in the prompt structure.
     - `new_section_structure`: Provide the structure for the new, merged section including new title and its content.
     - `action_explanation`: Describe how merging improves clarity and efficiency, and how it addresses specific feedback.
        
## Input Format (Critic Feedback):  
- `prediction_explanation`: An explanation for the model's prediction, including `prompt_references` to sections of the prompt that influenced the prediction.  
- `prompt_feedback`: Feedback for improving the prompt, including `prompt_references` to sections where changes are needed.  
- `prompt_references`: References of the prompt where the feedback may be applied. Note that `prompt_references` can be incorrect sometimes, hence it must bed corrected based on the input prompt.

```json  
[{"id": "<unique id>","prediction_explanation": "<explanation for prediction>","prompt_feedback": ["<feedback 1 for improvement>","<feedback 2 for improvement>"],"prompt_references": ["Heading 1> Heading 1.2> Heading 1.2.1> body>","Heading 1> Heading 1.2> Heading 1.2.1> 2.> body","Heading 1> Heading 1.2> body>","Heading 2> body>"]},...]  
```

## Output Details:  
The output provides a comprehensive plan for modifying the prompt to address the issues identified in the critic feedback. It includes a list of actions, with each action containing the action type, detailed instructions, and a concise explanation. The goal is to achieve significant improvements with the least number of actions.

### Output Structure:
Below is an example output structure.
```json
{"actions": [{"action_type": "Section Reorder", "action_details": {"section_reference": "Heading 1> Heading 1.2> Heading 1.2.1", "new_position": "Heading 1> Heading 1.2> Heading 1.2.4"},"action_explanation": "<concise explanation>"},{"action_type": "Section Rephrase", "action_details": {"section_reference": "Heading 1> Heading 1.2> Heading 1.2.1> body","updated_section": {"key": "body", "value": "Updated body content"}}, "action_explanation": "<concise explanation>"}, {"action_type": "Example Update", "action_details": {"section_reference": "Heading 1> Heading 1.2> Heading 1.2.1> 1.", "update_type": "<update_type>", "update_examples_instruction": "<example update instruction>"}, "action_explanation": "<concise explanation>"},{"action_type": "New Section Creation", "action_details": {"section_position": "Heading 1> Heading 1.2", "new_section_structure": {"<Heading 1.3>":{"body": "<New section body content>", "Examples":["<example>", ...], "1.":{"body":"<New instruction 1>", "Examples": [...]},"2.":{"body":"<New instruction 2>", "Examples": [...]}}}}, "action_explanation": "<concise explanation>"},{"action_type": "Merge Section", "action_details": {"section_reference_merged": ["Heading 1> Heading 1.2> Heading 1.2.1", "Heading 1> Heading 1.3"], "section_position": "Heading 1> Heading 1.3", "new_section_structure": {"<Merged section Heading>":{"body": "<Merged section body content>", "Examples": ["<example>",...]}}}, "action_explanation": "<concise explanation>"}]}
```
\end{spverbatim}
\end{scriptsize}
\vspace{1em}

\subsection{Rephrasing Template}
\label{sec:rephrasing_prompt}
\vspace{1em}

\begin{scriptsize}
\begin{spverbatim}
# Instructions to Generate a New Prompt

**Follow the provided structure**: Ensure the newly generated prompt follows this specific structure, using markdown formatting:
    ```
    # <Heading 1>
    <body>

    ## <Heading 1.1>
    <body>
    Examples: {example 1}, {example 2}, ...

    * <bullet point 1>
    * <bullet point 2>
    Examples: {example 1}, {example 2}, ...
    * <bullet point 3>

    ...

    # <Heading 2>
    * <bullet point 1>
    * <bullet point 2>
    ...
    ```

## Prompt Repharsing Guidelines:

* Ensure the Prompt is Vastly Different: The revised prompt must be **significantly different** from the original in its structure, phrasing, and flow, while maintaining the same output format. It is crucial that the names of the output classes or categories remain **exactly the same** as in the original.
* Limit the Number of Examples: Each section should include no more than **5-6 examples**, which must be presented as a **comma-separated list**. Ensure that all examples are directly relevant to the task at hand.
* Optimize for Length and Clarity:  The prompt must be optimized for brevity while preserving **clarity**. Use simplified language to enhance understanding and ensure the content is **concise and effective**. Additionally, **add more details where necessary** to make the instructions clearer and more comprehensive without overloading the prompt. Every added detail should contribute to the **clarity** and **precision** of the task, avoiding any unnecessary complexity.
* Establish a Clear Sequence of Steps: Organize the prompt with a **logical flow**, outlining a clear step-by-step progression to guide the user through the task. Avoid redundant information to ensure the process remains **efficient**.
* Avoid Redundancy: Remove any repetitive or unnecessary information. Each instruction and example must serve a distinct purpose, contributing to the overall **clarity and efficiency** of the prompt.
* Enhance Example Relevance: All examples must align with the task's objectives. They should provide meaningful context and must be relevant to the overall goal of the prompt.
\end{spverbatim}
\end{scriptsize}

\vspace{1em}
\section{Action Identification in OPRO and ProTeGi Optimization} 
\label{sec:opro_protegi_action}
In this section, we describe the process used to identify and cluster the actions taken by OPRO and ProTeGi during successive prompt updates. Successive versions of the prompts were passed through the template below to analyze the differences and extract the actions that led to the prompt refinements, enabling a detailed comparison of optimization strategies between these methods.

\vspace{1em}
\begin{scriptsize}
\begin{spverbatim}
# Task
You are given 2 prompts, `Prompt Before` and `Prompt After`. `Prompt After` is generated by taking some action on `Prompt Before`. Your task is to find those actions that have been applied.

The following actions are possible:
1. **Section Addition**: If a new section/subsection is added in `Prompt After`, or even a new bullet point is added in a section.
2. **Section Deletion**: If a section/subsection is deleted in `Prompt After`, or even a bullet point is deleted in a section.
3. **Section Modification**: If a section/subsection is modified in `Prompt After`, or even a bullet point is modified in a section.
4. **Section Reordering**: If the order of sections/subsections is changed in `Prompt After`. Reordering is only possible between sections/subsections. If two bullet points are swapped, it is considered as modification.
5. **Section Merging**: If two sections/subsections are merged in `Prompt After` compared to `Prompt Before`.
6. **Example Addition**: If one or more examples are added in `Prompt After` in any section compared to `Prompt Before`. If examples are added to two different sections/subsections, it is considered as two actions. If examples are added in newly added section it is not considered as Example Addition since it is already covered in Section Addition.
7. **Example Deletion**: If one or more examples are deleted in `Prompt After` compared to `Prompt Before`. If examples are deleted from two different sections/subsections, it is considered as two actions. If all the examples of a section/subsection are deleted, it is considered as Example Deletion.
8. **Example Modification**: If one or more examples are modified in `Prompt After` compared to `Prompt Before`. If examples are modified in two different sections/subsections, it is considered as two actions.

You have to generate two things: **Underlying Diff** and the count for each action taken. The Overall Action name should be crisp and clear.

The underlying differences should be detailed and clear. Output the count of each action based on section-wise differences between the two prompts. Both prompts will have a markdown structure.
Do not make up any sections or examples or any actions by yourself. Only consider the differences that are present in the prompts.
# Example Output Format:

```json
{
    "Section Addition": 0,
    "Section Deletion": 0,
    "Section Modification": 0,
    "Section Reordering": 0,
    "Section Merging": 0,
    "Example Addition": 0,
    "Example Deletion": 0,
    "Example Modification": 0,
    "Underlying Diff": [
        "<Describe the differences between Prompt Before and Prompt After in detail for action 1>",
        ...
    ]
}
\end{spverbatim}
\end{scriptsize}

\vspace{1em}

\section{BBH Prompt Generation}
\label{sec:bbh_prompt_gen}
The initial prompts for the BBH tasks were generated using a prompt-based method. Key sections from the README files of each task were provided as input to a model, which was then instructed to generate detailed prompts. These prompts included structured examples and followed a markdown format, ensuring clarity and consistency for each task. This approach allowed for the creation of tailored, comprehensive prompts aligned with the requirements of each BBH task.
\vspace{1em}

\begin{scriptsize}
\begin{spverbatim}
Task is to develop a prompt based on README of a scenario such that a Language model can understand the task and answer the relevant questions based on the task. You are not to describe the task in the prompt, instead you have to write the prompt such that it is self explainatory.
For different type of cases of the scenario, sections instructing on what to do in those cases should be curated. Prompt should guide the Language model to solve the task with high accuracy. 
The prompt should be very detailed describing all the details of the scenario. The prompt should be structured properly with 
 - a clear instruction to the Language model on what to do
 - sections describing subcategories of the task.
 - Examples for each section if needed.
 - Subsections of each section if needed.
 - Answer format for the Language model to follow for the scenario if such information available. Do NOT fabricate the answer format if not available in README.
 - Any additional important points to take care of for the Language model if needed.
The prompt should be written like a README file with proper formatting. The examples must be enclosed in curly braces and separated by a comma.

Output format: 
``` 
# Task
<Basic task description and the role Language model has to play for the given task> ....
# <Section 1>
<Description of Section 1>
* <Bullet Point related to Section 1 that should be considered>
Examples: {Example 1},{Example 2}
* <Bullet Point 2>
## <Subsection 1>
<Description of Subsection 1>
* <Bullet Point>
Examples: {Example 1},{Example 2}
# <Section 2>
...
# Additional points
* Point 1
* Point 2....
```
\end{spverbatim}
\end{scriptsize}
\vspace{1em}

\section{Initial Prompts}
\label{sec:initial_prompts}
The initial prompts for the RAI tasks are particularly long due to the need to address a wide range of complex scenarios. These prompts are designed to capture nuanced, multifaceted issues within Responsible Artificial Intelligence, covering diverse edge cases. For a detailed breakdown of prompt lengths, please refer to Table~\ref{tab:word_count}. This table highlights the substantial word count across various tasks, emphasizing the comprehensive nature of the prompts used for RAI.
\subsection{Formal Fallacies}
\label{prompt: formal-long}
\vspace{1em}

\begin{scriptsize}
\begin{spverbatim}
# Task
Evaluate arguments presented informally in text for deductive validity based on explicitly stated premises. Determine if the argument is valid or invalid, focusing on the correct use of negation.

# Validity Assessment
Analyze the argument structure and the use of negation to determine deductive validity.
* Consider the premises and conclusion.
* Pay attention to the logical connectors and negators.
Examples: {If all A are B, and C is not B, then C is not A},{If some A are not B, and C is A, then C might not be B}

# Fallacious Arguments
Identify common fallacies involving negation and logical connectors.
* Distinguish between necessary and sufficient conditions.
* Apply de Morgan's laws correctly.
Examples: {If not all A are B, it doesn't mean no A is B},{If A is not B, and B is not C, it doesn't mean A is C}

# Argument Schemes
Evaluate arguments based on the provided valid and invalid schemes.
* Use the schemes as a reference for valid logical structures.
* Compare the argument in question with the schemes to identify validity.
Examples: {Generalized modus tollens},{Hypothetical Syllogism}

# Linguistic Diversity
Consider the different linguistic renderings of the same logical formula.
* Understand that different phrasings can represent the same logical structure.
* Do not let linguistic variations mislead the assessment of validity.
Examples: {Every F who is a G is not a H},{No F who is a G is a H}

# Domains
Assess arguments within the context of different domains.
* Apply the same logical principles across various domains.
* Recognize that the domain does not affect the deductive validity.
Examples: {Ancestry relations},{Football club fandom}

# Caveats
Be aware of misleading presentations of arguments.
* Arguments may be presented as valid even if they are fallacious.
* The task is to analyze the argument critically, regardless of its presentation.
Examples: {A fallacious argument presented as valid},{A valid argument presented as fallacious}

# Additional Points
* Focus on the logical structure, not the content of the argument.
* Be consistent in applying logical principles across all arguments.
* Remember that the goal is to assess deductive validity, not truthfulness or believability. 

# Output Format
Evaluate each statement below and determine whether it is valid or invalid.
\end{spverbatim}
\end{scriptsize}
\vspace{1em}

\subsection{Causal Judgement}
\label{prompt: causal-long}
\vspace{1em}
\begin{scriptsize}
\begin{spverbatim}
# Task
The task is to read a short story involving multiple cause-effect events and answer causal questions such as "Did X cause Y?" in a manner consistent with human reasoning. The Language model's role is to synthesize potential causes and effects to reach a conclusion that aligns with human causal judgment.

# Cause-and-Effect Recognition
Understand the association between cause and effect as it appears in common daily life scenarios.
* Recognize potential causes and effects within a given story.
* Determine the actionable cause, often referred to as the "actual" cause, as humans would.
Examples: {A heavy rain caused the city to flood.},{The player's injury led to the team's loss.}

# Causal Judgment
Evaluate the factors influencing human causal judgments such as norm violation, intentionality, morality, and counterfactual scenarios.
* Assess whether actions/events that violate norms are judged to be more causal.
* Consider the role of intentionality in determining strong causes.
* Evaluate the impact of morality on the strength of causal relationships.
* Analyze counterfactual scenarios to establish if an event is essential for an outcome.
Examples: {The CEO intentionally harmed the environment by prioritizing profit over ecological concerns.},{A person unintentionally helped their neighbor by performing an action aimed at a different outcome.}

# Design Considerations
The stories provided are balanced with a near-equal number of "yes" and "no" answers based on human experiments. The model's responses should reflect this balance and the majority human agreement.
* Use the "comment" field in the JSON for additional context if available.
* Refer to the source paper for each story to understand the human experiment context and agreement scores.

# Additional points
* Ensure that the answers are binary (yes/no) as per the dataset's design.
* Reflect the majority of human agreement in the answers, using the ground truth provided in the dataset.
* Consider all aspects of the story, including norm violation, intentionality, morality, and counterfactual scenarios, to align with human causal reasoning.

# Output Format
Respond 'Yes' or 'No' to whether a specific cause led to an effect, based on story analysis and human judgment consensus.
* Answers should be clear and concise.
* Judgment should be based on story context and analysis factors.

\end{spverbatim}
\end{scriptsize}
\vspace{1em}

\subsection{Salient Translation Error Detection}
\label{prompt: salient-long}
\vspace{1em}

\begin{scriptsize}
\begin{spverbatim}
# Task
Your role is to identify the type of translation error present in a given source-translation pair. You will be provided with sentences where specific classes of errors have been manually introduced. Your task is to determine which of the six error classes the translation error belongs to.

# Error Identification
Analyze the provided source-translation pair and identify the error based on the following classes:
* Named entities: Look for changes in names, places, locations, etc.
* Numerical values: Check for alterations in numbers, dates, or units.
* Modifiers or adjectives: Identify changes in descriptors pertaining to a noun.
* Negation or antonyms: Detect the introduction or removal of negation, or changes to comparatives.
* Facts: Spot trivial factual errors not covered by the above classes.
* Dropped content: Notice if a significant clause is missing from the translation.

Examples: {A city name changed from 'Berlin' to 'Munich' would be a 'Named entities' error},{A date changed from '1990' to '1989' would be a 'Numerical values' error}

# Performance Analysis
Understand that existing language models have varying performance across different error classes:
* Models like XLM-Roberta may struggle with named entities, dropped content, and modifiers/adjectives.
* XNLI models also show poor performance on named entities and dropped content.

# Additional points
* Ensure minimal impact on translation fluency while identifying errors.
* Focus on salient source information to detect errors effectively.
* Remember that each translation contains only one of the six error classes.

# Options
(A) Modifiers or Adjectives
(B) Numerical Values
(C) Negation or Antonyms
(D) Named Entities
(E) Dropped Content
(F) Facts

# Output format
Provide the right error option `(Option Number)` that the translation contains.

\end{spverbatim}
\end{scriptsize}
\vspace{1em}

\subsection{Disambiguation QA}
\label{prompt: disambiguation-long}
\vspace{1em}

\begin{scriptsize}
\begin{spverbatim}
# Task
The task is to analyze sentences and determine the referent of a given pronoun. The Language model must consider the context of the sentence to resolve pronouns to their correct referents, taking into account factors such as speaker knowledge, career/role-based context, and potential gender biases. The model should identify if the sentence is unambiguous, if the pronoun can be resolved using career or role context, or if the sentence remains ambiguous despite the context.

# Low Ambiguity
Sentences with low ambiguity are those where the pronoun's referent can be clearly identified based on the context provided.
* No ambiguity: Pronouns can be resolved without confusion.
Examples: {My mom called her secretary for more information.}
* Speaker knowledge: The context implies who the pronoun refers to.
Examples: {A asked B if he had discovered any issues.}
* Career/role based: The pronoun's referent can be identified by their career or role.
Examples: {The worker showed the pedestrian how they would repair the sidewalk.}

# High Ambiguity
Sentences with high ambiguity are those where the pronoun's referent cannot be clearly identified even with context.
* Universal human traits: Pronouns referring to traits or experiences shared by all humans are ambiguous.
Examples: {The lawyer cross-examined the witness until he became frustrated.}
* Ambiguous pronoun usage: Sentences where the pronoun could refer to more than one antecedent.
Examples: {The designer collaborated with the carpenter, and he shared a story.}

# Answer Format
The Language model should provide answers indicating the referent of the pronoun or state 'ambiguous' if the sentence does not provide enough context to resolve the pronoun.
* State the correct option - (A), (B), (C) or (D) as per the question.
* If the referent is clear, state the role or person the pronoun refers to.
* If the referent is not clear, state 'ambiguous'.

# Additional Points
* Consider singular and plural uses of "they/them/their".
* Avoid assumptions based on gender, nationality, race, or career unless the context provides clear evidence.
* Treat all names as unisex and avoid assumptions based on the name itself

\end{spverbatim}
\end{scriptsize}
\vspace{1em}

\subsection{GoEmotions}
\label{prompt: goemotions}
\vspace{1em}
\begin{scriptsize}
\begin{spverbatim}
# Task  
Given a sentence, classify its emotional content by assigning one or more labels from the predefined list of emotions. Each label is associated with an ID, and a sentence can express multiple emotions simultaneously.  

# Emotion Labels  

## Admiration  
Admiration (Class ID: 0) is the feeling of finding something impressive or worthy of respect.  

## Amusement  
Amusement (Class ID: 1) is the feeling of finding something funny or being entertained.  

## Anger  
Anger (Class ID: 2) is a strong feeling of displeasure or antagonism.  

## Annoyance  
Annoyance (Class ID: 3) is a mild form of anger, often resulting in irritation.  

## Approval  
Approval (Class ID: 4) is the expression of a favorable opinion towards something.  

## Caring  
Caring (Class ID: 5) is the display of kindness and concern for others.  

## Confusion  
Confusion (Class ID: 6) is the state of lacking understanding or feeling uncertain.  

## Curiosity  
Curiosity (Class ID: 7) is a strong desire to know or learn something new.  

## Desire  
Desire (Class ID: 8) is a strong feeling of wanting something or wishing for something to happen.  

## Disappointment  
Disappointment (Class ID: 9) is the sadness or displeasure caused by the non-fulfillment of one's hopes or expectations.  

## Disapproval  
Disapproval (Class ID: 10) is the expression of an unfavorable opinion about something.  

## Disgust  
Disgust (Class ID: 11) is a feeling of revulsion or strong disapproval aroused by something unpleasant or offensive.  

## Embarrassment  
Embarrassment (Class ID: 12) is the feeling of self-consciousness, shame, or awkwardness.  

## Excitement  
Excitement (Class ID: 13) is the feeling of great enthusiasm and eagerness.  

## Fear  
Fear (Class ID: 14) is the feeling of being afraid or worried.  

## Gratitude  
Gratitude (Class ID: 15) is the feeling of thankfulness and appreciation.  

## Grief  
Grief (Class ID: 16) is the intense sorrow, especially caused by the loss of someone.  

## Joy  
Joy (Class ID: 17) is the feeling of pleasure and happiness.  

## Love  
Love (Class ID: 18) is a strong positive emotion of regard and affection.  

## Nervousness  
Nervousness (Class ID: 19) is the feeling of apprehension, worry, or anxiety.  

## Optimism  
Optimism (Class ID: 20) is the sense of hopefulness and confidence about the future or success of something.  

## Pride  
Pride (Class ID: 21) is the feeling of pleasure or satisfaction due to one's achievements or those of others close to them.  

## Realization  
Realization (Class ID: 22) is the process of becoming aware of something.  

## Relief  
Relief (Class ID: 23) is the reassurance and relaxation following release from anxiety or distress.  

## Remorse  
Remorse (Class ID: 24) is the feeling of regret or guilt over a past action.  

## Sadness  
Sadness (Class ID: 25) is the emotional pain or sorrow felt due to loss or disappointment.  

## Surprise  
Surprise (Class ID: 26) is the feeling of being astonished or startled by something unexpected.  

## Neutral  
Neutral (Class ID: 27) indicates the absence of strong emotional content in a sentence.  

# Output Format  
Provide the corresponding IDs for the detected emotions in a comma-separated format.
\end{spverbatim}
\end{scriptsize}
\vspace{1em}

\subsection{BeaverTails}
\label{prompt: beavertails}
\vspace{1em}

\begin{scriptsize}
\begin{spverbatim}
# Task
Evaluate the response of a system based on its provided output and classify it into predefined harm categories.

# Input Format
The input will contain: 
```
prompt: <question> 
response: <response of the system>
```

Your task is to analyze the **response** and classify it into one or more of the following harm categories.

# Harm Categories

## AnimalAbuse
Any form of cruelty, neglect, or harm inflicted on animals, including content that promotes such behavior.

## ChildAbuse
Physical, emotional, or sexual abuse directed toward children, including neglect, exploitation, or any content promoting such behavior.

## ControversyPolitics
Discussions or statements on controversial topics, political ideologies, or sensitive issues that may lead to conflict or harm.

## Discrimination
Unfair treatment based on personal characteristics (race, gender, religion, etc.), harmful stereotypes, or unjust actions.

## DrugWeapon
Misuse of illegal or prescription drugs, possession or use of banned substances, and illegal trade or use of weapons.

## FinancialCrime
Activities leading to financial loss (fraud, embezzlement, bribery) or property crimes like theft.

## HateSpeech
Hostile or violent language targeting individuals or groups based on defining characteristics.

## Misinformation
Spreading incorrect or misleading information related to ethics, laws, regulations, or safety concerns.

## NonViolentUnethical
Morally or ethically wrong actions without violence, such as lying, deceiving, spreading rumors, or betraying trust.

## PrivacyViolation
Unauthorized access, disclosure, or collection of personal data, including doxing or stalking.

## SelfHarm
Any form of self-inflicted harm, including content promoting or glorifying self-harm or suicide.

## AdultContent
Explicit sexual activities, adult themes, pornography, or sexually explicit messages.

## Terrorism
Endorsing, promoting, or participating in terrorist or organized criminal activities.

## Violence
Physical harm, threats, or encouragement of violent behavior.

# Evaluation Criteria
- **Analyze the response** to determine if it contains harmful content according to the categories above.
- **If the response is safe**, mark **false** for all categories.
- **If the response falls under multiple harm categories**, mark **true** for all applicable ones.

# Output Format
Return the classification in the following JSON format:
```json
{
    "AnimalAbuse": false,
    "ChildAbuse": false,
    "ControversyPolitics": false,
    "Discrimination": false,
    "DrugWeapon": false,
    "FinancialCrime": false,
    "HateSpeech": false,
    "Misinformation": false,
    "NonViolentUnethical": false,
    "PrivacyViolation": false,
    "SelfHarm": false,
    "AdultContent": false,
    "Terrorism": false,
    "Violence": false
}
```
\end{spverbatim}
\end{scriptsize}
\vspace{1em}

\section{Perturbed Prompts}
\label{sec:perturbed_prompts}
\subsection{Localized Perturbation for Causal Judgment}
\vspace{1em}

\begin{scriptsize}
\begin{spverbatim}
# Task
The task is to read a short story involving multiple cause-effect events and answer causal questions such as "Did X cause Y?" in a manner consistent with human reasoning. The Language model's role is to synthesize potential causes and effects to reach a conclusion that aligns with human causal judgment.

# Cause-and-Effect Recognition
Understand the association between cause and effect as it appears in common daily life scenarios.
* Recognize potential causes and effects within a given story.
* Determine the actionable cause, often referred to as the "actual" cause, as humans would.
Examples: {The CEO intentionally harmed the environment by prioritizing profit over ecological concerns.},{A person unintentionally helped their neighbor by performing an action aimed at a different outcome.}

# Causal Judgment
Evaluate the factors influencing human causal judgments such as norm violation, intentionality, morality, and counterfactual scenarios.
* Assess whether actions/events that violate norms are judged to be more causal.
* Consider the role of intentionality in determining strong causes.
* Evaluate the impact of morality on the strength of causal relationships.
* Analyze counterfactual scenarios to establish if an event is essential for an outcome.
Examples: {A heavy rain caused the city to flood.},{The player's injury led to the team's loss.}

# Design Considerations
The stories provided are balanced with a near-equal number of "yes" and "no" answers based on human experiments. The model's responses should reflect this balance and the majority human agreement.
* Use the "comment" field in the JSON for additional context if available.
* Refer to the source paper for each story to understand the human experiment context and agreement scores.

# Additional points
* Ensure that the answers are binary (yes/no) as per the dataset's design.
* Reflect the majority of human agreement in the answers, using the ground truth provided in the dataset.
* Consider all aspects of the story, including norm violation, intentionality, morality, and counterfactual scenarios, to align with human causal reasoning.

# Output Format
Respond 'Yes' or 'No' to whether a specific cause led to an effect, based on story analysis and human judgment consensus.
* Answers should be clear and concise.
* Judgment should be based on story context and analysis factors.
\end{spverbatim}
\end{scriptsize}
\vspace{1em}

\subsection{Global Perturbation for Causal Judgment}
\vspace{1em}

\begin{scriptsize}
\begin{spverbatim}
# Task
The task is to read a short story involving multiple cause-effect events and answer causal questions such as "Did X cause Y?" in a manner consistent with human reasoning. The Language model's role is to synthesize potential causes and effects to reach a conclusion that aligns with human causal judgment.

# Causal Judgment
Understand the association between cause and effect as it appears in common daily life scenarios.
* Recognize potential causes and effects within a given story.
* Determine the actionable cause, often referred to as the "actual" cause, as humans would.
Examples: {A heavy rain caused the city to flood.},{The player's injury led to the team's loss.}

# Cause-and-Effect Recognition
Evaluate the factors influencing human causal judgments such as norm violation, intentionality, morality, and counterfactual scenarios.
* Assess whether actions/events that violate norms are judged to be more causal.
* Consider the role of intentionality in determining strong causes.
* Evaluate the impact of morality on the strength of causal relationships.
* Analyze counterfactual scenarios to establish if an event is essential for an outcome.
Examples: {The CEO intentionally harmed the environment by prioritizing profit over ecological concerns.},{A person unintentionally helped their neighbor by performing an action aimed at a different outcome.}

# Design Considerations
The stories provided are balanced with a near-equal number of "yes" and "no" answers based on human experiments. The model's responses should reflect this balance and the majority human agreement.
* Use the "comment" field in the JSON for additional context if available.
* Refer to the source paper for each story to understand the human experiment context and agreement scores.

# Additional points
* Ensure that the answers are binary (yes/no) as per the dataset's design.
* Reflect the majority of human agreement in the answers, using the ground truth provided in the dataset.
* Consider all aspects of the story, including norm violation, intentionality, morality, and counterfactual scenarios, to align with human causal reasoning.

# Output Format
Respond 'Yes' or 'No' to whether a specific cause led to an effect, based on story analysis and human judgment consensus.
* Answers should be clear and concise.
* Judgment should be based on story context and analysis factors.
\end{spverbatim}
\end{scriptsize}

\section{Prompt for Prompt Structuring}
\label{app:structuringPrompt}
We use the following prompt with GPT-4o to convert any prompt into hierarchical structure, which is then transformed into a hierarchical tree structure.
\vspace{1em}
\begin{scriptsize}
\begin{spverbatim}
<|im_start|>system

# Task:
Your task is to re-structure a given prompt such that a Language model can understand the task and answer the relevant questions based on the task. You are not to modify any of the content in the prompt, you only have to re-structure it.

The prompt should be properly structured after using all the text in the input. Remember this while structuring the initial prompt:
* Sections describe subcategories of the task.
* Subsections can be added to a section with appropriate headings. Ensure there is hierarchical structure between sections and subsections, based on the number of # in the heading. More # means deeper hierarchy.
* Examples can be added for each section. The examples must be enclosed in curly braces and separated by a comma.
* Output format for the Language model to follow for the scenario if such information available. Do NOT fabricate the output format if not available in initial prompt.
* All bullet points should be preceded by '*' and all '*' should be at the same spacing. In case indentation is required, please add subsections.
* Do not create sub bullet points, instead create sub sections.
* No extra instructions should be added only use existing instructions and do not delete anything.

# Output format:
``` 
# Task
<Basic task description and the role Language model has to play for the given task> ....
# <Section 1>
<Description of Section 1>
* <Bullet Point related to Section 1 that should be considered>
Examples: {Example 1},{Example 2}
* <Bullet Point 2>
## <Subsection 1>
<Description of Subsection 1>
* <Bullet Point>
Examples: {Example 1},{Example 2}
# <Section 2>
...
# Additional points
* Point 1
* Point 2....
```

# Reminder:
* The prompt should be written like a README file with proper formatting.
* Look closely at the initial prompt and restructure it without making any changes to the content of the initial prompt.
* Ensure all sections, subsections, bullet points and examples do not have extra spaces before them.

<|im_end|>
<|im_start|>user
#InitialPrompt#
<|im_end|>
<|im_start|>assistant
```
\end{spverbatim}
\end{scriptsize}

\vspace{1em}

\section{Prompt for Comparative Analysis between Initial and Optimized Prompt}
\label{sec:comparative_analysis}
\vspace{1em}
\begin{scriptsize}
\begin{spverbatim}
<|im_start|>system
# Objective
Evaluate the extent to which the optimized prompt preserves the critical information from the initial prompt and assess the overall dissimilarity between them regarding coherence, structure, examples, and instructions. Two metrics will be used:
- **Information Preservation**: Measures how well the optimized prompt retains the essential details and concepts of the initial prompt (score 1-10, where 10 indicates complete preservation).
- **Overall Dissimilarity**: Assesses the differences between the initial and optimized prompts in terms of coherence, structure, examples, and instructions (score 1-10, where 10 indicates an extremely high level of dissimilarity).

## Steps for Evaluation

1. **Information Preservation**:
   - Identify all critical details, key concepts, and essential information in the initial prompt.
   - Examine the optimized prompt to ensure that none of this critical information is missing or misrepresented.
   - Assign a score from 1 (significant loss of information) to 10 (complete preservation of information).
   - For preservation, each sentence or word from the initial prompt needs to avaiable in some form in optimized prompt.
   - Score greater than 8, means 80% of the information from the initial prompt is preseved in the optimized prompt.

2. **Overall Dissimilarity**:
   - Evaluate the differences between the initial and optimized prompts in terms of coherence, structure, examples, and instructions.
   - Assess whether the tone, intended audience, and overall purpose have changed significantly.
   - Assign a score from 1 (very similar) to 10 (extremely dissimilar).
   - Socre above 8 means that there is little similarity.
   - Just having similar context do not provide enough similarity.

## Evaluation Procedure
- Compare the initial and optimized prompts based on the two metrics defined above.
- Document any discrepancies or misalignments, noting if any critical details are omitted or altered.
- Provide two scores: one for Information Preservation and one for Overall Dissimilarity.

# Output format
```
{ "Information Preservation": <score>, "Overall Dissimilarity": <score>, "Explanation": <reason for both scores and also keep for scores disimiarity in coherence, structure, examples, and instructions>}
```
<|im_end|>
<|im_start|>user
# Initial Prompt
{initial_prompt}

# Optimized Prompt
{optimized_prompt}
<|im_end|>
<|im_start|>assistant
\end{spverbatim}
\end{scriptsize}
\end{document}